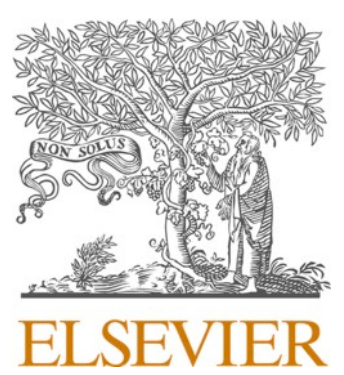

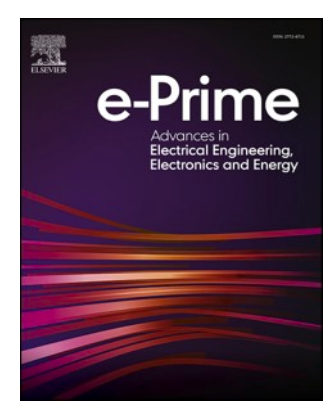

# An adaptive multi-fuzzy logic model for diagnosing transformer faults using dynamic weight optimization

Kim-Anh Nguyen [*], Huy Hoang Le, Ba Tu Phung

*Faculty of Electrical Engineering, the University of Danang - University of Science and Technology, Da Nang, Vietnam*

ARTICLE INFO

*Keywords:*
Power transformer fault diagnosis
Dissolved gas analysis
Multi-fuzzy logic
Adaptive weight adjustment
Dynamic optimization

ABSTRACT

Dissolved gas analysis (DGA) is crucial for diagnosing early power transformer failures. Traditional DGA interpretation methods like Duval Triangle, IEC ratio, Roger ratio, Doernenburg ratio and Key Gas are inconsistent and vary in accuracy, especially for multiple fault conditions. We propose an Adaptive Multi-Fuzzy Logic (AMFL) model integrating multiple DGA methods with fuzzy logic and a dynamic weight adjustment mechanism. Unlike existing approaches with fixed weights, this system iteratively evaluates each method's diagnostic performance, identifies multiple fault types, and adjusts weights based on fault prediction accuracy. A feedback-based optimization recalibrates weights after each cycle to ensure optimal solution convergence. The model, implemented in MATLAB/Simulink, is validated against DGA datasets with known error conditions. Results show the AMFL model significantly improves diagnostic accuracy, especially in complex error scenarios, and enhances adaptability to new datasets. Comparative analysis demonstrates the proposed method outperforms traditional fixed weight multi-fuzzy systems in accuracy, consistency, and reliability of error detection. This work provides a robust, flexible diagnostic tool for transformer condition monitoring and supports more accurate asset management decisions.

*Abbreviations*

| | |
|---|---|
| ADASYN | Adaptive Synthetic Sampling |
| AMFL | Adaptive Multi-Fuzzy Logic |
| ANFIS | Adaptive Neuro-Fuzzy Inference System |
| ANN | Artificial Neural Network |
| ARLT | Association Rules Learning Technique |
| BWOA | Binary Whale Optimization Algorithm |
| CNN | Convolutional Neural Network |
| DBN | Deep Belief Network |
| DGA | Dissolved Gas Analysis |
| DL | Deep Learning |
| DPFD | Data Preprocessed Fuzzy Logic Duval Triangle Method |
| DPM | Duval Pentagon Method |
| DRM | Doernenburg Ratio Method |
| DST | Dempster–Shafer Theory |
| DTM | Duval Triangle 1 Method |
| ENN | Edited Nearest Neighbor |
| FIM | Fault Interpretation Matrix |
| FIS | Fuzzy Inference System |
| FL | Fuzzy Logic |
| FL-DGA | Fuzzy Logic – Dissolved Gas Analysis |
| FL-DRM | Fuzzy Logic – Doernenburg's Ratio Method |
| FL-DTM | Fuzzy Logic – Duval's Triangle Method |
| FL-IRM | Fuzzy Logic – IEC Ratio Method |
| FL-KGM | Fuzzy Logic – Key Gas Method |
| FL-RRM | Fuzzy Logic – Roger's Ratio Method |
| FLTFDS | Fuzzy Logic Transformer Fault Diagnostic System |
| GA | Genetic Algorithm |
| IEC | International Electrotechnical Commission |
| IRM | IEC Ratio Method |
| KGM | Key Gas Method |
| K-MCA | K-Means Clustering Algorithm |
| KNN | K-Nearest Neighbors |
| LSSVM | Least-Squares Support Vector Machine |
| MLP | Multi-Layer Perceptron |
| MSE | Mean Squared Error |
| PSO | Particle Swarm Optimization |
| QML | Quantum Machine Learning |
| RBF | Radial Basis Function |

* Corresponding author.
*E-mail address:* nkanh@dut.udn.vn (K.-A. Nguyen).

| | |
|---|---|
| RL | Reinforcement Learning |
| RRM | Roger's Ratio Method |
| SVFB | Support Vector Feature-Based |
| SVM | Support Vector Machine |
| TLBO | Teaching-Learning Based Optimization |
| VQSL | Variational Quantum Shadow Learning |
| D | Discharge |
| D1 | Discharges of Low Energy |
| D2 | Discharges of High Energy |
| DT | Combined Thermal and Discharge |
| PD | Partial Discharge |
| T | Thermal Fault |
| UD | Undefined Diagnosis |
| $CH_4$ | Methane |
| $C_2H_6$ | Ethane |
| $C_2H_4$ | Ethylene |
| $C_2H_2$ | Acetylene |
| CO | Carbon Monoxide |
| $CO_2$ | Carbon Dioxide |
| $H_2$ | Hydrogen |

## 1. Introduction

Power transformers are fundamental to the reliability and stability of modern electrical power systems, serving as critical nodes in energy transmission and distribution networks. Any unexpected failure in transformers can result in costly disruptions, unplanned outages, and operational hazards [1,2]. Condition monitoring and fault diagnostics, therefore, play an indispensable role in minimizing risks and ensuring timely maintenance [3,4]. Among various diagnostic techniques, DGA has become the industry standard for detecting early-stage faults in transformers [5]. By analyzing the concentrations of gases dissolved in transformer oil, such as hydrogen ($H_2$), methane ($CH_4$), ethane ($C_2H_6$), ethylene ($C_2H_4$), acetylene ($C_2H_2$), and carbon monoxide (CO), DGA provides a reliable means of identifying thermal faults, electrical discharges, and insulation degradation [6,7]. Traditional methods for interpreting DGA, including Key Gas analysis, ratio and graphical methods, have been widely employed to associate gas patterns with fault types [8–11]. However, these approaches face limitations in diagnosing cases with overlapping fault characteristics or gas ratios that fall outside established thresholds [6].

Over the past few years, many smart techniques have been proposed to overcome the drawbacks of traditional DGA techniques in power transformer fault diagnosis. Initial attempts were being made in utilizing Artificial Neural Networks (ANNs) due to their rule extraction and classification properties. For example, Bhalla *et al.* [12] used ANN for human-interpretable diagnosis rule extraction, whereas Seifeddine *et al.* [13] used Multi-Layer Perceptron (MLP) and Radial Basis Function (RBF) networks with better performance compared to conventional methods on low-dimensional data sets. Then, Chatterjee *et al.* [14] integrated an MLP with fuzzy boundaries into the Duval Pentagon 1 model for enhancing decision precision in boundary regions. In the same way, Ghoneim *et al.* [15] integrated different DGA interpretation methods into one ANN for generalization improvement using various gas patterns.

Adaptive Neuro-Fuzzy Inference Systems (ANFIS) is a hybrid method integrating the learning capability of neural networks and reasoning ability of fuzzy systems. ANFIS was also optimized in [16] with Binary Whale Optimization Algorithm (BWOA) and association rules learning technique (ARLT) for improved feature extraction. Khan *et al.* [17] showed that reinforced ANFIS outperformed standalone fuzzy systems, whereas Kari *et al.* [18] combined ANFIS with Dempster–Shafer theory for the improvement of diagnostic certainty via evidence fusion.

Support Vector Machines (SVMs) also exhibited good performance, particularly when combined with optimization algorithms. Benmahamed *et al.* [19] combined SVM with Particle Swarm Optimization (PSO) and k-Nearest Neighbors (KNN) for the Duval Pentagon 1 model to enhance classification accuracy. Further improvements were also observed in [20], in which a hierarchical decision tree was combined with Support Vector Feature-Based (SVFB) to tackle rigid class boundaries. In [21], Differential Evolution and Grey Wolf Optimizer were fused with Least-Squares Support Vector Machine (LSSVM) for removing local minima and enhancing fault classification reliability.

In data imbalance handling, Rajesh *et al.* [22] contrasted the impact of Adaptive Synthetic Sampling (ADASYN) method enhancing the generalization of the machine learning model considerably on 4,580 field-collected DGA samples. Azmi *et al.* [23] utilized Edited Nearest Neighbor (ENN) with SVM and decision trees with 88% classification accuracy. Nanfak *et al.* [24] extended the hybrid modeling concept with evolutionary k-means clustering and expert systems with 98.29% accuracy on complicated fault classes.

The advent of Deep Learning (DL) witnessed a dramatic enhancement. Taha *et al.* [25] employed Convolutional Neural Networks (CNNs) in proposing a sturdy model with ±20% noise tolerance in gas input concentrations. Jin *et al.* [26] also employed CNNs on online DGA data streams to carry out real-time transformer health predictions with 87% accuracy. Ghoneim *et al.* [27] used Teaching-Learning-Based Optimization (TLBO) for dynamic threshold tuning, whereas Zou [28] suggested a Deep Belief Network (DBN) model with flexible input gas combinations and stable multi-class classification.

In Reinforcement Learning (RL), Malik *et al.* [29] proposed a fuzzy reinforcement learning-based classifier for DGA-based diagnosis. The proposed model employed the J48 decision tree algorithm for feature selection and achieved 99.7% accuracy, revealing the tremendous potential of adaptive fuzzy agents for power transformer monitoring.

Most recently, Quantum Machine Learning (QML) has been investigated as a next-generation approach. He *et al.* [30] proposed a revised Variational Quantum Shadow Learning (VQSL) framework using localized variational circuits to learn quantum shadow features. This approach achieved 95.4% diagnostic accuracy and verified the viability of quantum-enhanced transformer fault classification.

Among them, Fuzzy Logic (FL) is still fundamental because it has the intrinsic capability of handling fuzzy boundaries and uncertainty in gas ratios. Nguyen *et al.* [31] employed FL on both the International Electrotechnical Commission (IEC) and Roger key gas methods to surmount strict ratio constraints. In another article that year, FL was coupled with a strong data preprocessing phase to enhance the diagnostic precision of the Duval Triangle 1 method from 90% to 97.1% accuracy [32]. Other research, including [33,34], and [35], also demonstrated the efficacy of FL for integrating multiple DGA indicators. Abu-Siada *et al.* [1] particularly developed an FL-backed system that integrated several DGA methods and obtained better classification accuracy through ensemble reasoning.

Building on these foundations, this study proposes an Advanced Multi-Fuzzy Logic (AMFL) framework that enhances both the flexibility and diagnostic power of traditional FL-based systems. Specifically, the FL interpretations of five classical DGA methods, IEC Ratio (FL-IRM) [36,37], Roger's Ratio (FL-RRM) [38–40], Doernenburg's Ratio (FL-DRM) [41], Duval's Triangle 1 (FL-DTM) [42,43], and Key Gas Method (FL-KGM) [44], have been upgraded through redefinition of membership functions and rule enhancement, improving fault detection sensitivity and granularity.

A key innovation of the AMFL model lies in the dynamic weight adaptation mechanism, which automatically adjusts the influence of each FL-DGA method based on its diagnostic accuracy. This allows the model to emphasize more reliable methods for specific fault cases and improves its adaptability to diverse datasets. Furthermore, the fuzzy membership functions are deliberately designed with overlapping regions, allowing the system to make soft inferences in uncertain or boundary conditions—something traditional methods struggle with. This is particularly beneficial for detecting overlapping fault signatures and rare or ambiguous fault types.

In addition, AMFL is capable of handling "Undefined Diagnosis" (UD) outcomes from individual methods without failing entirely. The ensemble structure ensures that as long as at least one method produces a valid result, the overall system remains functional. The model also extends the output space to include mixed faults—especially DT (simultaneous thermal and electrical)—by combining partial outcomes across multiple methods. This capacity significantly enhances its real-world diagnostic utility, where complex fault scenarios often arise.

Finally, the AMFL model exhibits strong generalization performance without the need for balanced or large-scale training datasets. By preserving the natural distribution of both training and testing data, the model maintains high diagnostic accuracy (> 99%) while remaining robust and scalable in practical applications. These strengths collectively make AMFL a powerful, interpretable, and industry-ready solution for transformer fault diagnosis.

The remainder of this paper is organized as follows. Section 2 introduces the proposed AMFL model, detailing the enhancements made to each individual fuzzy logic-based DGA interpretation method, including the redefinition of membership functions and the expansion of fuzzy rule bases. This section also describes the architecture of the weight adjustment mechanism, which iteratively recalibrates the contribution of each diagnostic method based on its historical accuracy. Section 3 presents the dataset used for model training and testing, the preprocessing techniques applied to the raw DGA data, and the experimental setup implemented in MATLAB/Simulink. This section also includes a comprehensive performance evaluation of the AMFL model compared to traditional and fixed-weight multi-FL approaches, highlighting improvements in diagnostic accuracy and adaptability across diverse fault scenarios. Finally, Section 4 summarizes the key contributions of the study and discusses future directions, including the integration of quantum machine learning for enhancing real-time diagnostic capability and extending the framework to alternative transformer insulation systems.

## 2. Methodological background

This section describes the methodology underlying the proposed AMFL model, focusing on the optimization of FL-DGA methods and incorporation of a dynamic weight adjustment mechanism. These innovations aim to address the limitations of traditional FL-systems and improve diagnostic accuracy for complex fault scenarios in power transformers.

### 2.1. Traditional DGA interpretation methods

DGA is widely recognized as the industry standard for transformer fault diagnostics, due to its capability to identify incipient faults before they escalate into catastrophic failures. Through the analysis of gas concentrations dissolved in transformer insulating oil, DGA provides critical insights into the internal conditions of transformers, facilitating effective fault detection and timely maintenance interventions [1,4,5]. Key diagnostic gases are indicative of specific fault mechanisms, such as thermal degradation (T) characterized by $C_2H_6$ and $C_2H_4$, electrical discharges (D) identified by $C_2H_2$, or insulation breakdowns. Consequently, DGA serves as an invaluable tool for condition monitoring [9].

The efficacy of DGA is attributed not only to its capacity to detect a diverse range of fault conditions but also to the diagnostic methodologies employed to interpret gas concentrations [6]. Conventional approaches, including Key Gas Method (KGM) [6], Doerenburg's Ratio Method (DRM), IEC Ratio Method (IRM) [9], Roger's Ratio Method (RRM) [10], Duval's Triangle 1 Method (DTM) [7,11], have been extensively utilized to correlate characteristic gas patterns with specific fault types. For instance, the DTM demonstrates particular effectiveness in diagnosing electrical anomalies, such as voltaic arcs, while the KGM is frequently applied to identify thermal faults [4]. Notwithstanding their widespread application, these methods possess inherent limitations [5]. One particularly notable limitation is observed in the DRM, which requires two initial conditions to be satisfied before proceeding with fault diagnosis. These conditions involve validating specific gas ratios and thresholds, which, if not met, result in a "non-diagnosable" state [6,9]. Moreover, the static diagnostic rules inherent in these methods constrain their adaptability to varying operational conditions or complex fault scenarios, such as concurrent DT fault [8].

In addition to detecting faults in oil, DGA also evaluates the degradation of paper insulation. Cellulosic thermal decomposition produces CO and carbon dioxide ($CO_2$) at lower temperatures than oil decomposition, and traceable amounts of these gases can be detected even under normal operating conditions [3]. To make a confirmed decision on the condition of paper degradation, the $CO_2$/CO ratio is utilized [9,45]. This ratio provides valuable insights into the extent of insulation aging or thermal stress, complementing other diagnostic techniques in DGA.

### 2.2. FL-system overview

FL-systems have emerged as promising solutions to the limitations of conventional DGA interpretation methods, offering a robust framework for handling uncertainties and ambiguities in gas analysis data [33]. Unlike traditional techniques with rigid thresholds, FL employs flexible membership functions that classify data into overlapping categories, enabling more effective fault classification under uncertain conditions [34,35]. This adaptability is particularly valuable for managing ambiguities in DGA datasets, where gas concentration levels may not align precisely with predefined fault categories.

The initial implementation of the multi-FL approach was introduced in the study by Q. Su [37], which pioneered the integration of multiple DGA methods within a FL-framework for transformer fault diagnosis. Building upon this foundation, Abu-Siada further developed and applied the approach in subsequent research [1]. Additionally, other studies, such as [39], utilize multi-FL-DGA methods and a FL layer as a decision-making mechanism, and [40], which employs a Fault Interpretation Matrix (FIM), have also adopted multi-FL-DGA techniques. These studies reinforce the effectiveness of integrating multiple diagnostic methods to enhance fault classification accuracy. However, their static membership functions and limited FL-rules configurations limit their adaptabilities to novel datasets and complex fault scenarios. By overcoming the constraints of static configurations, the proposed model offers improved flexibility and performance across diverse datasets and fault complexities. In contrast, the proposed FL-system incorporates:

#### 2.2.1. Optimized membership functions

The refinement of output membership functions for fault classification was conducted to enhance the sensitivity and precision of transformer fault diagnosis. A primary challenge of conventional FL approaches in DGA is the use of predefined membership functions, which may not accurately represent the complex gas concentration patterns associated with various fault types. These limitations stem from rigid boundaries between fault categories, complicating the diagnosis of cases with gas concentrations near transition regions. To address these shortcomings, the newly designed membership functions, particularly for the FL-IRM, as depicted in Fig. 1, were optimized through statistical analysis of historical fault data [31]. The adjustment process involved examining a comprehensive dataset of transformer faults, identifying gas concentration trends, and redefining membership functions to better capture subtle variations in gas composition. By incorporating a data-driven approach, the improved membership functions now offer smoother transitions between fault categories, thereby reducing the likelihood of misclassification. Furthermore, these refinements facilitate more accurate differentiation between complex fault scenarios, such as distinguishing between overheating and partial discharge when gas concentration levels overlap. Unlike traditional approaches that often assign a fault type based on a rigid rule set, the enhanced method provides a more flexible and adaptive classification process, allowing for a more precise diagnosis. Consequently, the new membership functions bridge the diagnostic gaps left by conventional DGA methods, significantly improving the reliability of transformer fault detection and classification. The remaining membership functions for other FL-based DGA methods will be provided in Appendix A.

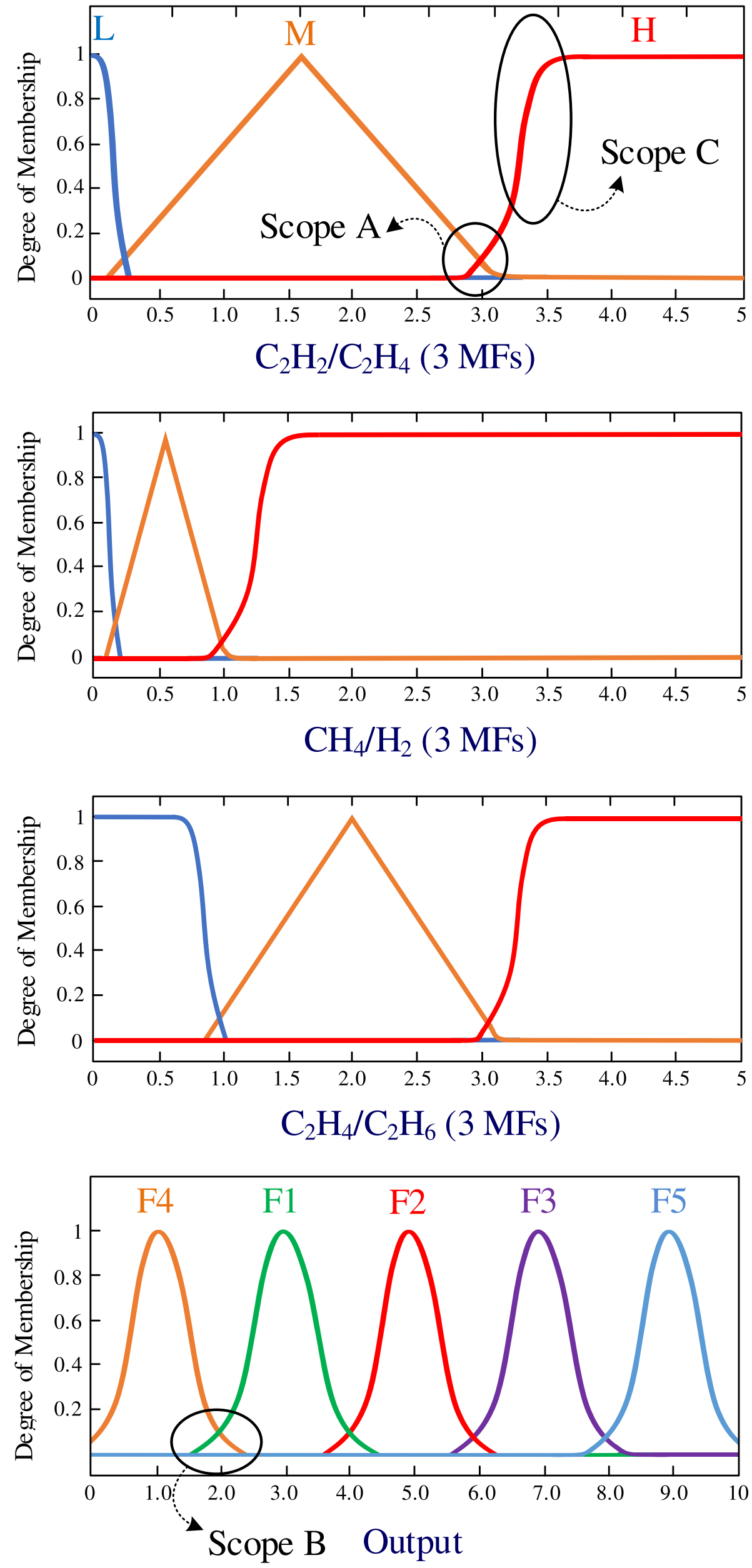

**Fig. 1.** Redesigned membership functions for the FL-IRM.

The redesign of membership functions (MFs) in our fuzzy logic DGA modules was guided by three key improvements aimed at enhancing diagnostic performance and boundary handling capabilities, as illustrated in Fig. 1:

- Intentional overlapping between adjacent MFs (e.g., Scope A and Scope B): One fundamental feature of fuzzy logic is its ability to handle uncertainty through partial membership. In overlapping regions like Scopes A and B, input values belong to two fuzzy sets simultaneously (e.g., 30% in set X and 70% in set Y). This design contrasts with rigid threshold-based methods by offering a soft transition, enabling the system to manage boundary cases more effectively. Combined with fuzzy rule sets and centroid-based defuzzification, this probabilistic handling improves diagnostic flexibility. Unfortunately, some prior works (e.g., [1,38,39]) neglected this principle and treated fuzzy boundaries as rigid thresholds, resulting in performance similar to traditional rule-based methods in edge cases.
- Shape and Symmetry of MFs (e.g., Scopes A, B, and C): In overlapping areas like Scopes A and B, symmetric triangular or trapezoidal functions are used to avoid biasing the result toward one region over another. In contrast, in isolated regions like Scope C, we intentionally apply sharply nonlinear functions to quickly amplify the membership value, making classification more decisive where no ambiguity exists.
- Range Selection and Statistical Refinement: Initially, fuzzy region boundaries were designed based on IEC standard thresholds (e.g., Table 1), with membership codes 0, 1, 2 representing Low (L), Medium (M), and High (H), respectively. However, this baseline design yielded limited accuracy. To improve this, we performed cluster analysis and distribution assessment on our dataset of 760 samples. Based on observed data trends and natural groupings, we adjusted the fuzzy boundaries to better match the actual behavior of fault-related gas ratios. For zones prone to overlap, we applied the overlapping strategy, ensuring smoother transitions and better classification robustness.

### 2.2.2. *Enhanced rule base*

To ensure robustness, multiple rules are incorporated into the FL-system to address specific fault cases in which traditional methods fail to diagnose them effectively. For instance, rule [0 1 1] has been added to FL-IRM's ruleset to address partial-discharge (PD) faults, which the traditional IRM method does not encompass. This topic will be explored in greater detail in Subsection 2.4.1.

### 2.2.3. *Defuzzification using the centroid method*

To converts FL outputs back into numerical values, the FL-system utilizes the centroid defuzzification method, which computes the center of gravity of the aggregated FL output [46]. This method is mathematically represented as:

$$D_{output} = \frac{\int \mu_{output}(x)xdx}{\int \mu_{output}(x)dx} \tag{1}$$

Where, $D_{output}$ is the defuzzified diagnostic output; $\mu_{output}(x)$ is the aggregated membership function for the output variable; $x$is the continuous range of possible output values. The centroid method ensures that the final diagnostic outcome represents the most balanced interpretation of all fuzzy rules, thereby mitigating bias and maintaining consistency in fault classification.

## 2.3. *Adaptive weight adjustment module*

An iterative mechanism recalibrates the weights of individual DGA methods, dynamically prioritizing their contributions based on diagnostic performance. This comprehensive approach improves transformer health assessment, potentially identifying faults that may be overlooked by single methods [1,37]. The system's adaptive nature, featuring membership functions and weight adjustment, facilitates continuous refinement of diagnostic capabilities as it processes additional data, ensuring that the model can respond to evolving operational conditions. The architecture of this proposed model is depicted in Fig. 2, which illustrates the integration of multiple FL modules, weight adaptation mechanisms, and iterative outputs to ensure accurate and dynamic fault classification.

**Table 1**
IEC ratio codes [9].

| Ratio | $C_2H_2/C_2H_4$ | $CH_4/H_2$ | $C_2H_4/C_2H_6$ |
|---|---|---|---|
| <0.1 | 0 | 1 | 0 |
| 0.1–1.0 | 1 | 0 | 0 |
| 1.0–3.0 | 1 | 2 | 1 |
| >3.0 | 2 | 2 | 2 |

Furthermore, the system demonstrates the capability to train with diverse datasets, allowing it to adapt and enhance its performance for specific faulty conditions. For instance, when trained on datasets predominantly comprising electrical faults, the system adapts by increasing the emphasis on methods that excel in electrical fault diagnosis, such as FL-DTM, while reducing the emphasis on thermally focused methods like FL-RRM or FL-KGM. Conversely, for training datasets emphasizing thermal faults, the system adjusts by prioritizing thermal-based approaches while downweighting those optimized for electrical fault detection. By leveraging historical data and advanced computational intelligence techniques, such as those highlighted in [47,48], the system addresses the limitations of traditional ratio-based methods [9–11]. These enhancements align with the advancements in transformer diagnostics discussed in prior studies [4,8], ensuring that the system evolves to meet the demands of complex and overlapping fault conditions.

## 3. Proposed adaptive multi-fuzzy logic model

### 3.1. Fuzzy logic-traditonal method

The proposed AMFL model incorporates two significant advancements to enhance diagnostic accuracy and address the limitations of traditional approaches. While retaining the foundational elements of the original framework as delineated in [1], the model incorporates significant enhancements to address more complex fault scenarios. Firstly, the FL framework for individual diagnostic methods (e.g., DTM, IRM, RRM, DRM, and KGM) has been significantly improved compared to the previous study [39,40,43,44]. These enhancements include the addition of several new rules to handle specific edge cases and improve the overall robustness of the system. For example, collections of rules have been developed to address conditions that traditional methods fail to diagnose, such as subtle variations in gas ratios or fault states that do not align with predefined boundaries [41]. These improvements enable the system to adapt more dynamically to diverse fault conditions and datasets, providing a more comprehensive and accurate diagnostic tool. Together, these advancements ensure that the enhanced FL model achieves superior performance in classifying even the most challenging fault scenarios.

In this study, the FL-IRM rules were formulated based on the original IRM code definitions outlined in Table 1 and Table 2, and subsequently

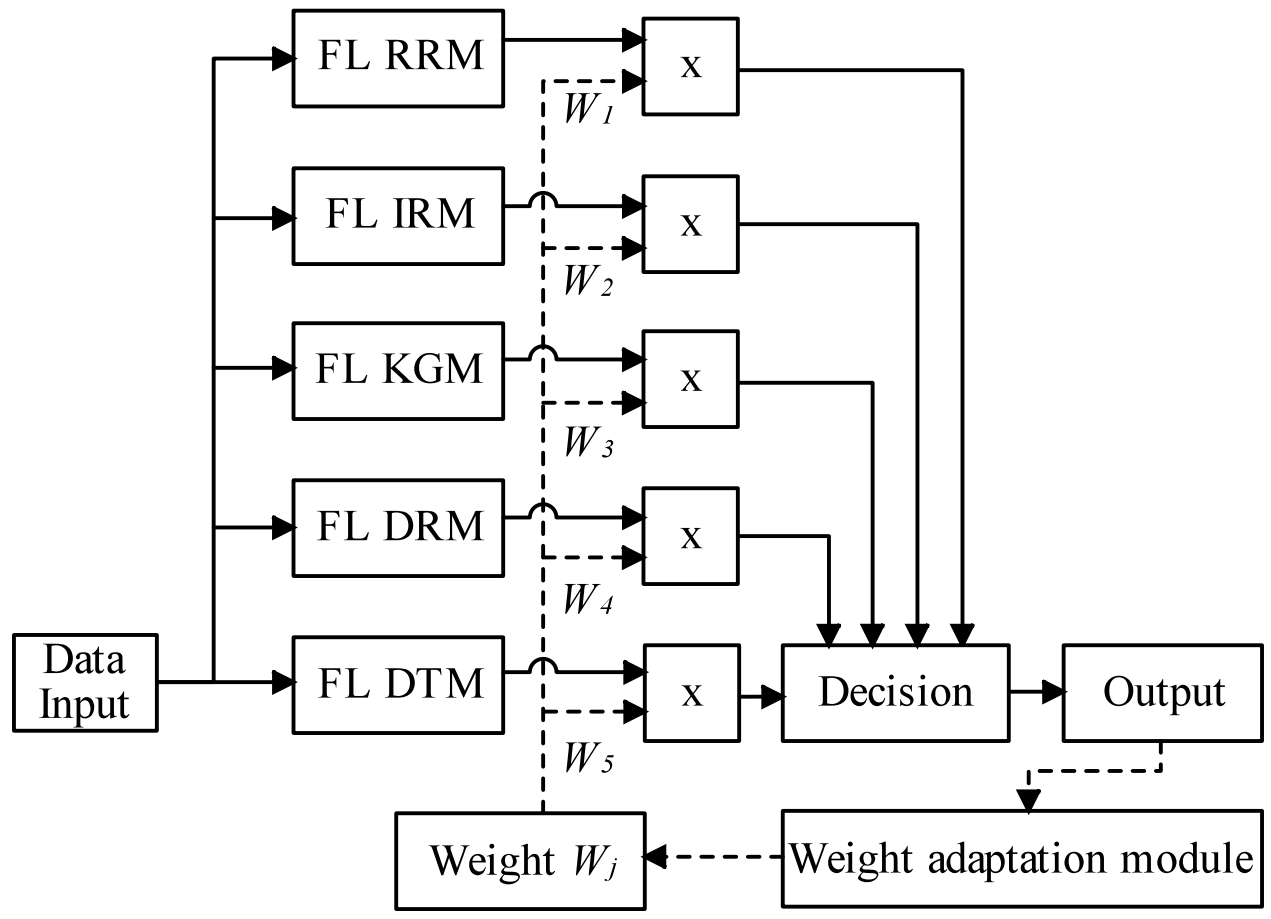

**Fig. 2.** Architecture of the proposed FL-based diagnostic model with adaptive weight mechanism.

**Table 2**
IEC codes to interpret incipient faults in transformers [9].

| Fault type | $C_2H_2/C_2H_4$ | $CH_4/H_2$ | $C_2H_4/C_2H_6$ |
|---|---|---|---|
| Normal deterioration | 0 | 0 | 0 |
| Partial discharges of low energy density | 0 | 1 | 0 |
| Partial discharges of high energy density | 1 | 1 | 0 |
| Discharges of low energy | 1 or 2 | 0 | 1 or 2 |
| Discharges of high energy | 1 | 0 | 2 |
| Thermal fault of temperature < 150 °C | 0 | 0 | 1 |
| Thermal fault of low temperature range 150°C to 300°C | 0 | 2 | 0 |
| Thermal fault of medium temperature range 300°C to 700°C | 0 | 2 | 1 |
| Thermal fault of high temperature range > 700°C | 0 | 2 | 2 |

The previous FL-IRM model contained a duplication of rule 1 and rule 17, with rule 1′s output failing to conform to the IEC code. This duplication resulted in the omission of a critical rule, [0 0 0], which corresponds to F5. In the original IEC framework, both rules 1 and 17 are associated with the IEC code [0 1 0] and should output PD (F2). However, in the prior study, rule 1 incorrectly produced F5 while rule 17 retained F2, causing inconsistencies. In this work, rule 1 has been corrected to output F2 (PD), and rule 17 has been replaced by the missing rule [0 0 0], ensuring proper alignment with the IEC standard.

augmented with six additional rules, as detailed in Table 3. Several inconsistencies found in previous FL-IRM models [1] have been corrected to improve diagnostic accuracy. One notable issue was the membership function for the $CH_4/H_2$ ratio, which deviated from the original IEC definitions. Specifically, IEC defines three ranges: Range 1 (0–0.1), Range 0 (0.1–1), and Range 3 (>1), as shown in Table 1. However, a prior study incorrectly extended Range 0 to 0.1–3, leading to inaccurate results. In this work, the membership function has been rectified to align with the original IEC methodology, as illustrated in Fig. 1. Additionally, errors in rule definitions have been addressed.

The fully revised and enhanced FL-IRM ruleset developed in this study is illustrated in Fig. 3. These figures collectively present the comprehensive modifications, including the updated membership functions, the corrected rule set, and the new rules introduced to handle additional fault scenarios, ensuring the robustness and accuracy of the diagnostic model.

Second, the decision-output framework of five FL-DGA methods has been expanded to incorporate a multi-fault state, DT, which addresses overlapping conditions where both thermal and electrical fault characteristics coexist. This enhancement is grounded in a detailed statistical analysis of historical DGA datasets, ensuring the model's capability to manage complex fault scenarios effectively. By redefining and refining the state intervals, the system reduces overlaps and ambiguity in fault classification while maintaining adherence to the foundational principles of the original design. The expanded decision-output was developed in accordance with Table 4, incorporating a new fault category, F6 (DT), to more effectively address concurrent thermal and electrical faults. To accurately detect DT faults, the system requires one or more methods to classify a fault as D and one or more methods to classify it as T. Consequently, the DT diagnostic value range was systematically designed as the means of the ranges where both D and T faults simultaneously manifest across multiple methods [6]. Specifically, in Table 4, the DT diagnostic value range is precisely divided into two principal ranges: 4.0–4.3, typically resulting from the occurrence of two methods classified as T and one classified as D, and 5.7–6.0, generally arising from two methods classified as D and one classified as T. These scopes were strategically developed based on observed fault characteristics and the methodology of DTM concerning the DT region, to ensure the most accurate and comprehensive assessment of transformer fault conditions. This refinement not only enhances diagnostic precision but also significantly increases the practicality of the method, ensuring it can effectively address real-world fault scenarios. The systematic placement of the DT intervals represents a substantial improvement over traditional approaches, underscoring the proposed system's capacity to offer a more nuanced and actionable evaluation of transformer health.

**Table 3**
Added rules for FL-IRM model.

| Ratio | | | Fault |
|---|---|---|---|
| $C_2H_2/C_2H_4$ | $CH_4/H_2$ | $C_2H_4/C_2H_6$ | |
| 0 | 1 | 1 | PD |
| 0 | 1 | 2 | PD |
| 2 | 1 | 1 | Discharge |
| 2 | 1 | 2 | Discharge |
| 1 | 2 | 0 | Heating |
| 0 | 0 | 2 | Heating |

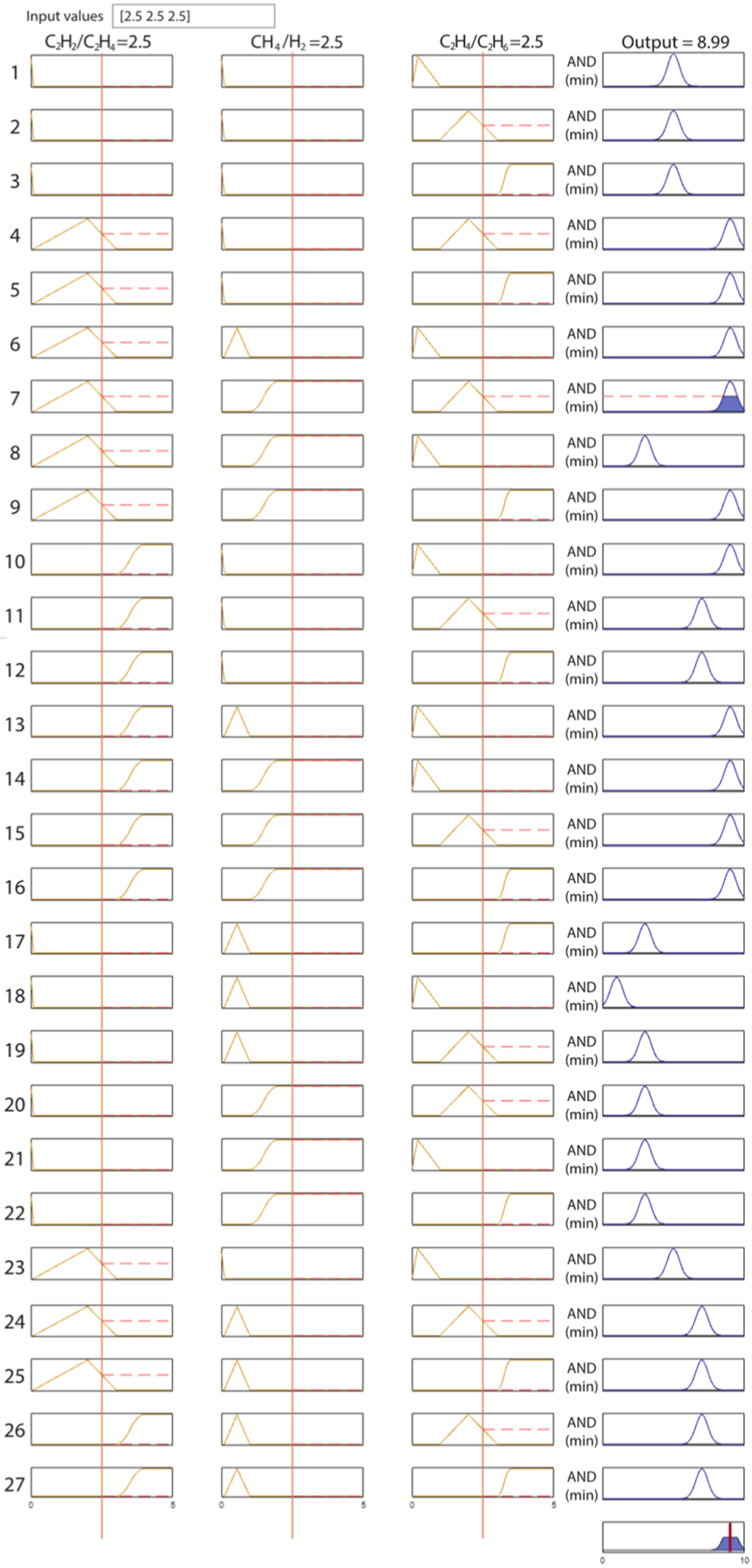

**Fig. 3.** Updated fuzzy ruleset for the FL-IRM.

**Table 4**
Fault code for overall output decision.

| Fault | Decision (model) | Sub-decision | Fault diagnosis |
|---|---|---|---|
| F4 | $0.0 < D \leq 2.0$ | | - Normal (N) |
| F1 | $2.0 < D \leq 4.0$ | $\frac{CO_2}{CO} \geq 7.0$ | - Localized overheating without cellulose involvement (T) |
| | | $\frac{CO_2}{CO} < 7.0$ | - Localized overheating with cellulose involvement (T & C) |
| F6 | $4.0 < D \leq 4.35.7 < D \leq 6.0$ | | - Mixed Thermal and Electrical (DT) |
| F2 | $4.3 < D \leq 5.7$ | | - Corona Partial Discharge in oil (PD) |
| F3 | $6.0 < D \leq 8.0$ | | - Discharges in oil (D) |
| F5 | $8.0 < D \leq 10.0$ | | - Out of Code |

The rulesets for the remaining diagnostic methods, including FL-DTM, FL-KGM, FL-RRM, FL-DRM, and other supplementary techniques, were designed and validated following a methodology consistent with that used for the FL-IRM approach presented earlier. Each method was enhanced through the refinement of membership functions and expansion of fuzzy rule bases, ensuring their adaptability and diagnostic accuracy across various fault scenarios. This unified design framework guarantees coherence among all FL-based modules within the AMFL system and contributes to its overall robustness. For instance, in our previous research [32], the accuracy of the FL-DTM was improved from 80% to 98.1% through the implementation of an enhanced preprocessing layer.

Building upon this advancement, the preprocessing layer not only enhances data quality but also extends the FL-DTM method by increasing the number of input gas ratios from three (%$CH_4$, %$C_2H_4$, and %$C_2H_2$) to five (%$H_2$, %$CH_4$, %$C_2H_6$, %$C_2H_4$, %$C_2H_2$). This improvement addresses a critical limitation of the traditional approach, wherein the absence of $C_2H_6$ compromises the accuracy of thermal fault diagnosis, and the exclusion of $H_2$ affects the detection of PD faults. For example, the classification results before and after employing the preprocessing layer for DTM, as illustrated in Fig. 4, demonstrate an increase in the model's accuracy. By incorporating these additional gas ratios, the refined model ensures a more comprehensive representation of transformer fault conditions, leading to improved classification performance and higher diagnostic reliability. This study also employs this enhancement by implementing the data preprocessed fuzzy logic-Duval triangle 1 method (DPFD), thereby further augmenting the system's reliability and robustness.

### 3.2. Adaptive multi-fuzzy logic intergrated

The proposed fault diagnosis framework, illustrated in Fig. 5, integrates five FL-DGA interpretation techniques into a comprehensive software model. This model is designed to assess transformer health and ascertain criticality levels based on DGA results. The KGM was initially employed due to its high accuracy in identifying normal operating conditions [1,6]. If the analysis indicates that the DGA data represents normal conditions, the system ceases further evaluation and reports the transformer as functioning within normal parameters [9]. However, in instances where the KGM detects abnormalities, subsequent analysis is conducted utilizing the five FL-DGA techniques.

Each technique evaluates the fault type and criticality of the transformer. For each iteration $k$, the combined diagnostic output $D_{combined,k}$ is calculated as a weighted sum of the outputs from all $N$diagnostic methods ($N = 5$). The formula is as follows:

$$D_{combined,k} = \frac{\sum_{j=1}^{N} W_j D_{ij}}{\sum_{j=1}^{N} W_j} \quad (2)$$

Where, $W_j$ represents the weight of method $j$, $D_{ij}$is the output of method $j$ for data point $i$. Equation (2) ensures that more accurate methodologies exert greater influence on the final determination, while less reliable approaches contribute proportionally less. Furthermore, any methodology yielding a diagnostic outcome inconsistent with anticipated fault patterns is excluded from the calculation by assigning its decision a value of zero. Subsequently, the adaptive weighting module employs training data and $D_{combined,k}$ to optimize the distribution of weights $W_j$. Thereafter, the combined diagnostic output $D_{combined,k}$ is recalculated based on the optimized weights $[W]$ and evaluated to classify the fault type and assess transformer criticality. In thermal fault scenarios (in which $2 < D_{combined,k} < 4$), the $CO_2$/CO ratio [45] is analyzed to differentiate between oil overheating and cellulose insulation deterioration. If the ratio suggests severe cellulose degradation ($CO_2/CO < 7$), further polymerization tests are recommended. Finally, the system either reports the identified fault type (F1, F2, F3 or F4) or confirms a normal condition if no significant fault is detected.

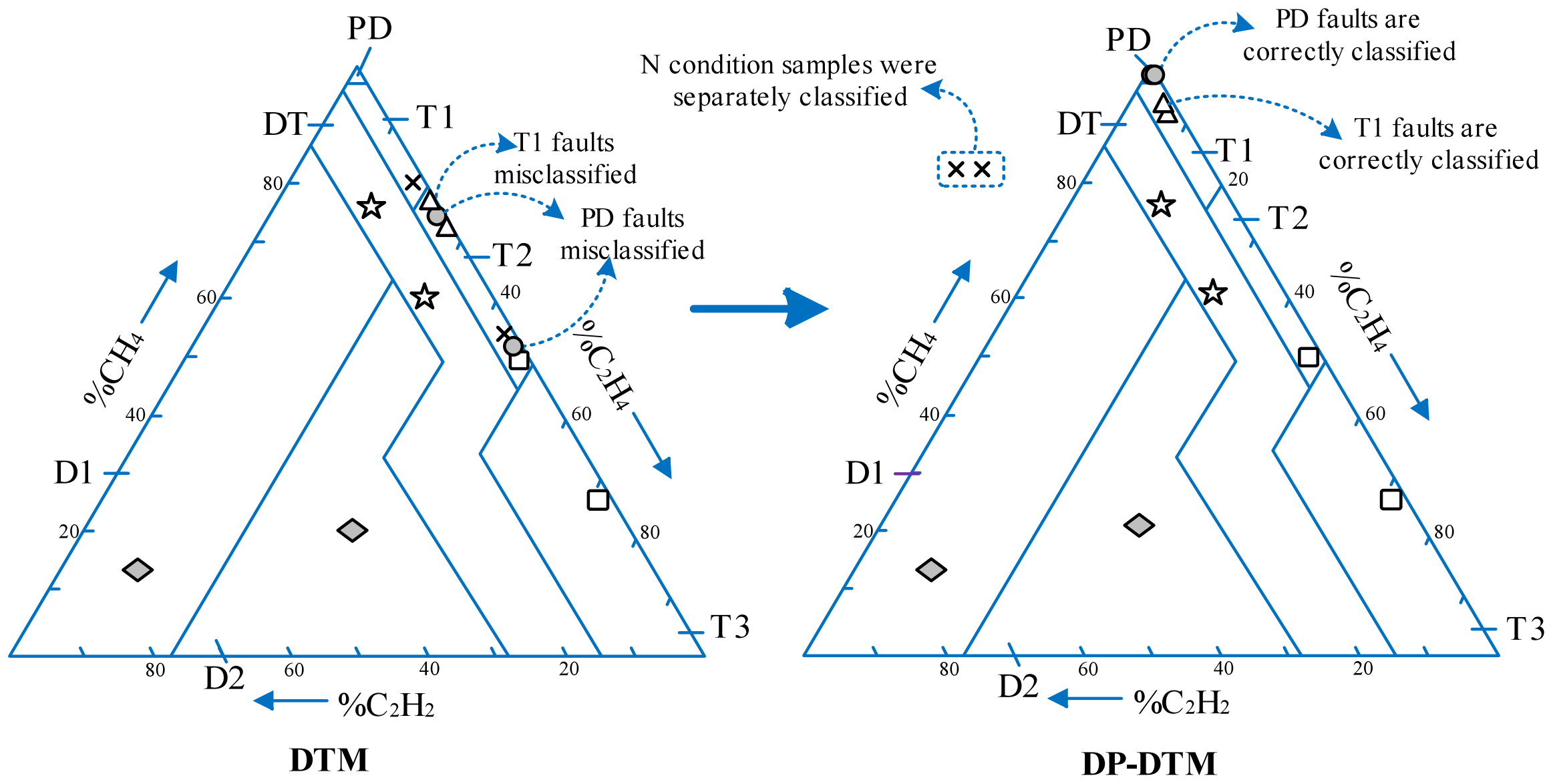

**Fig. 4.** Accuracy improvement in DTM after applying preprocessing layer [32].

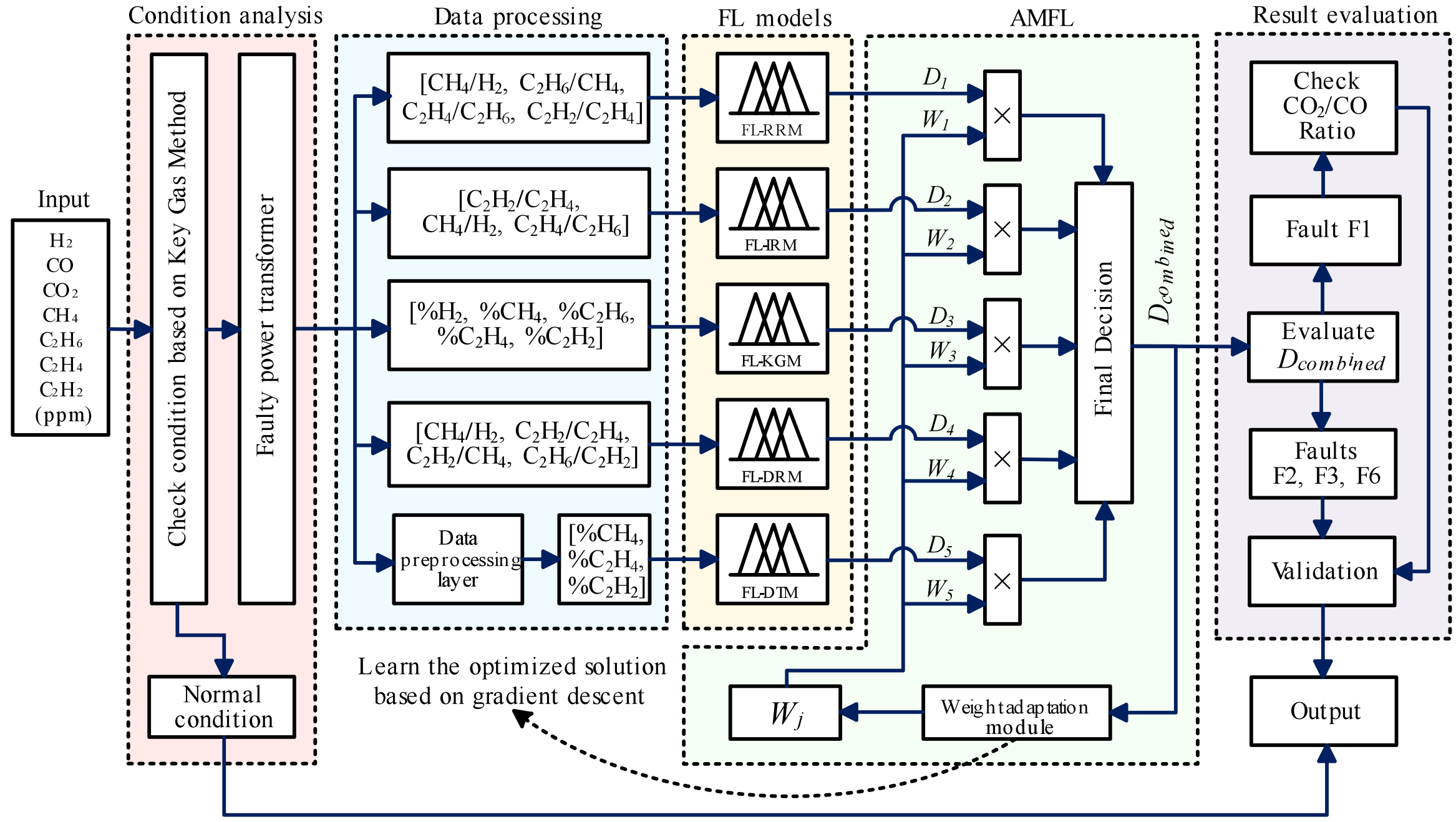

**Fig. 5.** Schematic representation of the system for diagnosing transformer faults.

### 3.2.1. *Proposed dynamic weight adjustment mechanism*

Static weights in traditional multi-FL-systems are inadequate for addressing dynamic variations in faulty conditions. Therefore, a weight adjustment mechanism is proposed. This mechanism is also called an adaptive mechanism and it operates by evaluating the performance of each diagnostic method after each dataset run. This iterative adjustment ensures that the system progressively prioritizes the most effective methods, enhancing diagnostic accuracy across diverse fault scenarios [47,48]. The proposed mechanism performs the following steps:

- **Step 1** – **Initialization**: The initial weights $W_{init}$ are assigned based on historical fault datasets and accuracy metrics obtained from prior studies [1]. These weights serve as the starting point for the iterative optimization process, ensuring a balanced contribution from all methods. These initial weights used in the adaptive optimization module were not randomly selected but were derived from the diagnostic accuracy of each FL-DGA method computed on the training dataset. This approach reflects the baseline performance of each method and provides a meaningful starting point for weight adaptation. A similar approach to initial weight assignment was also mentioned in [1], although it was not further explained and did not involve any adaptive mechanism. In contrast, our model incorporates an adjustment process that makes the initial weight selection even more critical to overall performance. Furthermore, the adaptive algorithm applies a constraint (denoted $\delta$) to limit the maximum allowable change in weight between runs. While this prevents overfitting and ensures convergence stability, it also restricts the optimization process from significantly altering weights that were initialized too far from their optimal values. For example, if the optimal weight for FL-IRM is 0.6 but the initial value is set at 0.2 and $\delta = 0.1$, the post-optimization range is limited to [0.1, 0.3], preventing convergence to the true optimal value. In contrast, initializing the weight closer to the optimum—say at 0.55 with $\delta = 0.1$—permits a range of [0.45, 0.65], which successfully encompasses the ideal value of 0.6. Therefore, by selecting initial weights based on observed performance and ensuring they lie near expected optimal regions, the model achieves better adaptability and higher final diagnostic accuracy. Across multiple experiments, we observed that using diagnostic accuracy–based initial weights consistently outperformed random initialization strategies, yielding better convergence and stronger final classification results. This validates our approach as a robust and effective initialization method for the adaptive multi-fuzzy logic framework.
- **Step 2** – **Error assessment**: The difference between the aggregated decision $D_{combined,k}$ and the observed true value $D_{real}$ for each data point is used to compute the squared error:

$$Error_k = \left(D_{combined,k} - D_{real}\right)^2 \tag{3}$$

- The total error for $M$ data points in $k-$th iteration, often referred to as the mean squared error $MSE_k$ in optimization contexts, is calculated as:

$$MSE_k = \frac{1}{M}\sum_{i=1}^{M} Error_k \tag{4}$$

- **Step 3** – **Weight optimization**: The weights $W_j$ are updated iteratively using a gradient descent approach to minimize the $MSE_k$. The process can be divided into the following three sub-steps:
- Gradient calculation: The gradient of the total error with respect to each $W_j$ is computed as:

$$\frac{\partial MSE_k}{\partial W_j} = -\frac{2}{M}\sum_{i=1}^{M}\left(D_{real} - D_{combined,k}\right)\left(\frac{\sum_j^N D_{ij}\left(\sum_j^N W_j\right) - \sum_j^N D_{ij}\, W_j}{\left(\sum_j^N W_j\right)^2}\right) \tag{5}$$

- Weight update rule: To minimize the *MSE*, the weights $W_j$ are updated iteratively using gradient-based optimization. The update formula for each weight $W_j{}^{(k+1)}$at $(k+1)-$th iteration is given by:

$$W_j^{(k+1)} = W_j^{(k)} - \eta \frac{\partial MSE_k}{\partial W_j} \tag{6}$$

- Where $\eta$ is the learning rate that controls the step size of updates.
- To prevent excessive changes in weights, a constraint $\delta$ is applied:

$$If \| W_{new} - W_{init} \|_1 > \delta then, scale W_{new} \tag{7}$$

- Where, $\| \ . \ \|_1$ represents the $L_1 -$ *norm* which measures the total magnitude of weight changes. The regularization step ensures that the optimization process remains stable and avoids overfitting specific data points.
- The algorithm imposes a constraint on the maximum allowable weight change using a vector-based normalization approach, as defined in Steps 10–14 of Algorithm 1. Rather than clipping each weight individually, the algorithm evaluates the L1-norm distance between the new weight vector $W_{new}$ and the initial vector $W_{init}$. If this distance exceeds a predefined threshold $\delta$, the weight update is rescaled proportionally along the direction of change, ensuring that the overall deviation remains within the constraint limit:

$$W = W_{init} + (W_{new} - W_{init})\frac{\delta}{\| W_{new} - W_{init} \|_1} \tag{8}$$

- This formulation preserves the directionality of the intended weight update while preventing excessive or unstable changes in a single iteration. Compared to simple range-based clipping (as used in the illustrative example in Step 1 of Subsection 3.2.1), this approach allows the optimization process to retain the full gradient information from the unconstrained update, rather than suppressing it arbitrarily. As a result, convergence becomes more stable, and the model is better able to adapt weights efficiently without overshooting or underreacting, especially in high-dimensional weight spaces.
- **Step 4** – **Convergence criteria:** The optimization process halts when one of the following conditions is met: *(i)* convergence the *MSE* below a specified threshold, indicating acceptable accuracy, or *(ii)* maximum iteration the algorithm completes the predefined number of iterations without achieving full convergence [49]. By adjusting the weights, this mechanism ensures that methods demonstrating higher reliability in specific fault scenarios are prioritized. This adaptability significantly enhances the system's diagnostic accuracy, particularly for complex and overlapping fault conditions.

### 3.2.2. *Weight optimization algorithm*

The Algorithm 1 is designed to iteratively refine weight values to minimize errors while preventing overfitting. It begins by initializing weight parameters and setting key hyperparameters, including a learning rate $\eta$ and a constraint $\delta$. During each iteration, the algorithm calculates the combined diagnostic output $D_{combined}$ as a weighted sum of diagnostic indicators, divided by the sum of weights. Subsequently, it determines the *MSE* between the computed and actual diagnostic values, followed by the gradient used to update the weights. After applying the gradient descent update, the weights are constrained to be non-negative and normalized to sum to one. If the new weight vector deviates significantly from the initial weights by more than $\delta$, it is adjusted to limit excessive changes, thereby ensuring stability and reducing overfitting. This process is repeated for a predefined number of iterations, refining the weight distribution for optimal performance. The complete procedure is detailed in Algorithm 1.

The following provides supplementary explanations for the initial parameters in Step 1. In particular,

- the weight vector is determined as follows: $W_{init} = [0.5 \quad 0.517 \quad 0.867 \quad 0.57 \quad 0.88]$. $W_{init}$ correspond to the initial weights of five FL-DGA modules, which come from the accuracy of those methods in study [1];
- $W = [W_1, \ W_2, \ W_3, \ W_4, \ W_5]$, where, each element in vector $W$corresponds to the initial weight assigned to one of the five FL-DGA modules: FL-RRM, FL-IRM, FL-KGM, FL-DRM, and FL-DTM, respectively;
- let $M_{tr}$ denote the number of training samples. The output matrix $D \in \mathbb{R}^{M_{tr}\times 5}$ contains the diagnostic results from the five FL-DGA modules for each training sample. Specifically: $D = [D_1, D_2, D_3, D_4, D_5]$, where each column $D_j$ corresponds to the output of module $j-$th for all training samples. The ground truth label vector $D_{real} \in \mathbb{R}^{M_{tr}\times 1}$ contains the actual fault types for each sample, encoded as continuous numerical values based on fault type: 1.00 for N, 3.00 for T, 4.15 for DT, 5.00 for PD, 7.00 for D. These values are directly labelled from the ground truth in the training dataset and based on the mean of each region in Table 4;
- other parameters include: $threshold = 10^{-5}$ is the convergence threshold for optimization; $max_iteration = 100$ is the maximum number of iterations allowed; $\eta = 0.1$ is the learning rate; $\delta = 2$is a constraint to ensure the weight change doesn't exceed the threshold to prevent overfitting; $1^T = [1 \quad 1 \quad 1 \quad 1 \quad 1]$ is an all-1 vector to broadcast row by row. The expression $sum(W) \times D - \left(D \times W^T\right) \times 1^T$in Step 6 produces a matrix of the same dimensions $[M_{tr} \times 5]$, ensuring that element-wise subtraction is valid. The term $D \times W^T$ results in a column vector $[M_{tr} \times 1]$, which is broadcast across all five columns via multiplication with $1^T = [1 \quad 1 \quad 1 \quad 1 \quad 1]$, yielding a matrix compatible for subtraction with $sum(W) \times D$.

To provide better clarity on the optimization mechanism behind Algorithm 1, the paper presents a detailed numerical example illustrating a complete iteration (e.g., iteration $k = 1$) in Appendix B. For simplicity, we chose a DGA dataset consisting of 6 samples. This example demonstrates how the adaptive weight adjustment is performed, starting from initialization to the weight update.

It is important to note that the learning rate $\eta$ and weight constraint threshold $\delta$ were empirically tuned to balance convergence speed and model stability. A value of $\eta = 0.1$ was chosen to ensure smooth yet responsive weight adaptation. The constraint parameter was set to $\delta = 2$, limiting excessive shifts in the weight vector per iteration while preserving the direction of optimization (as detailed in Algorithm 1). These values were found to provide optimal performance across repeated experiments.

## 4. Data processing, experiments and results

### 4.1. *Data description*

The dataset employed for this study contains DGA data used to assess transformer fault conditions. It provides a comprehensive record of gas concentrations $H_2$, $CH_4$, $C_2H_6$, $C_2H_4$, $C_2H_2$, CO and $CO_2$ in oil samples, along with corresponding fault classifications based on standards and expert analysis. Key characteristics of the dataset include:

**Algorithm 1**
Weight optimization algorithm.

Step 1: Initial Parameters
$W_{init} = [0.5, 0.517, 0.867, 0.57, 0.88]$, $W = [W_1, W_2, W_3, W_4, W_5]$;
$D = [D_1, D_2, D_3, D_4, D_5]$, $D_{real} = \begin{cases} 1.00 & \text{if } actual_fault = \text{N} \\ 3.00 & \text{if } actual_fault = \text{T} \\ 4.15 & \text{if } actual_fault = \text{DT} \\ 5.00 & \text{if } actual_fault = \text{PD} \\ 7.00 & \text{if } actual_fault = \text{D} \end{cases}$;
$threshold = 10^{-5}$, $max_iteration = 100$, $\eta = 0.1$; $\delta = 2$; $1^T = [1 \ \ 1 \ \ 1 \ \ 1 \ \ 1]$
Step 2: $W = W_{init}$
Step 3: **for** $k = 1$ to $max_iteration$ **do**
Step 4: $D_{combined,k} = \frac{D \times W^T}{sum(W)}$;
Step 5: $Error_k = (D_{combined,k} - D_{real})^2$;
Step 6: $gradient = -2\,mean\left((D_{real} - D_{combined,k})\ \frac{sum(W) \times D - (D \times W^T) \times 1^T}{(sum(W))^2}\right)$;
Step 7: $W_{new} = W - \eta \times gradient$;
Step 8: $W_{new} = \max(W_{new}, 0)$; % Ensure $W_{new}$ is greater than 0
Step 9: $W_{new} = \frac{W_{new}}{sum(W_{new})}$; % Normalize the total weight to 1
Step 10: **if** $\| W_{new} - W_{init} \|_1 > \delta$ **then**
Step 11: $W = W_{init} + (W_{new} - W_{init})\ \frac{\delta}{\| W_{new} - W_{init} \|_1}$; % Ensure the weight change does not exceed $\delta$
Step 12: **else**
Step 13: $W = W_{new}$;
Step 14: **end if**
Step 15: **end for**

- Source and features: The dataset consists of 760 data points collected from diverse and credible sources, including operational records from field transformers [50], and validated datasets from previous studies, as presented in Table 5. These data points span multiple years and encompass a wide range of dissolved gas measurements, with fault labels categorized into over heating oil (T), discharges of low energy (D1), discharges of high energy (D2), and PD. The diversity and comprehensiveness of the dataset ensure robustness and adaptability in fault diagnosis, enabling the proposed system to address complex and challenging diagnostic scenarios effectively.

- Data segmentation:
- Training data: 600 DGA data points used to optimize the AMFL model, including refining membership functions and adjusting weights;
- Testing data: 160 data points (43 data points from [50], 117 data points from IEC TC 10 databases [54]) reserved for validating the model's performance and assessing diagnostic accuracy.

This rich and diverse dataset enables the proposed system to adapt effectively to specific fault scenarios, such as PD, D or T, enhancing its reliability in real-world applications. Finally, Diagnostic accuracy is measured as the percentage of correctly classified faults. The accuracy $A$ of each methodology is determined by its efficacy in detecting faults as outlined below:

**Table 5**
Datasets source references and fault distribution of the 760 data points.

| Reference | Fault type | | | Total |
|---|---|---|---|---|
| | Thermal | Discharge | PD | |
| [13] | 13 | 7 | 1 | 21 |
| [21] | 7 | 4 | 7 | 18 |
| [50] | 201 | 248 | 52 | 501 |
| [51] | 25 | 19 | 7 | 51 |
| [52] | 6 | 3 | 1 | 10 |
| [53] | 21 | 13 | 8 | 42 |
| [54] | 34 | 74 | 9 | 117 |
| Total | 307 | 368 | 85 | 760 |

$$A = \frac{N_s}{N_t} \times 100\% \tag{9}$$

Where $A$ is the accuracy rate of caculated method, $N_s$ is the successful cases that the method predicted, and $N_t$ is total cases that we are testing. In this context, a successful case refers to an instance where the diagnostic method correctly identifies the fault type. Therefore, the accuracy metric reflects the proportion of correct predictions out of the total test cases. A higher accuracy value indicates better diagnostic performance, which is crucial in ensuring reliable fault identification for maintenance and operational decisions. This equation had also been used in many researches to evaluate the precision of their proposed models [1,23,55].

### 4.2. Data processing

Gas concentrations are normalized and mapped to FL input variables. The input data comprises the concentrations of seven distinct gases, while the output is represented as a real number ranging from 0.0 to 10.0, corresponding to specific fault conditions in the intervals defined in Table 4. These range from N to severe faults such as T, PD, D or DT faults.

Data is preprocessed before being fed into the FL-DGA model for fault detection. For ratio-based methods, the data is processed to produce ratios suitable for each method's input requirements. For instance, in the FL-IRM, the input comprises three gas ratios: $C_2H_2/C_2H_4$, $CH_4/H_2$, and $C_2H_4/C_2H_6$. Similarly, in the FL-RRM, the input consists of four gas ratios: $CH_4/H_2$, $C_2H_6/CH_4$, $C_2H_4/C_2H_6$, and $C_2H_2/C_2H_4$. For the FL-DRM, preprocessing involves two steps to validate the data before it is fed into the FL-block. These steps ensure valid input data based on four gas ratios: $CH_4/H_2$, $C_2H_2/C_2H_4$, $C_2H_2/CH_4$, and $C_2H_6/C_2H_2$.

For diagnostic methods such as FL-KGM and FL-DTM, the input data is no longer expressed in terms of gas ratios, but rather as the absolute percentages of individual gas components. Specifically, both the FL-KGM and the DPFD models utilize input variables consisting of $\%H_2$, $\%CH_4$, $\%C_2H_6$, $\%C_2H_4$, $\%C_2H_2$. To enhance overall readability, a summary mapping input gases to DGA methods and their corresponding fault classifications is provided in Table 6.

The overall data preprocessing stage has been illustrated in Fig. 6 and explained in more detail in Fig. 7, along with the fully developed AMFL model.

**Table 6**
Mapping input gases to DGA methods and their corresponding fault classifications.

| Input parameters | FL-DGA methods | Detected faults |
|---|---|---|
| $CH_4/H_2$, $C_2H_6/CH_4$, $C_2H_4/C_2H_6$, $C_2H_2/C_2H_4$ | FL-RRM | N, PD, T, D |
| $C_2H_2/C_2H_4$, $CH_4/H_2$, $C_2H_4/C_2H_6$ | FL-IRM | N, PD, T, D |
| $\%H_2$, $\%CH_4$, $\%C_2H_6$, $\%C_2H_4$, $\%C_2H_2$ | FL-KGM | N, PD, T, D |
| $CH_4/H_2$, $C_2H_2/C_2H_4$, $C_2H_2/CH_4$, $C_2H_6/C_2H_2$ | FL-DRM | N, PD, T, D |
| $\%H_2$, $\%CH_4$, $\%C_2H_6$, $\%C_2H_4$, $\%C_2H_2$ | DPFD | N, PD, T, D, DT |

### *4.3. MATLAB/Simulink-based simulation*

The simulation implementation of the proposed AMFL model using MATLAB/Simulink is presented in Figs. 6 and 7. These diagrams are derived from the schematic representation in Fig. 5, which outlines the overall architecture of the AMFL diagnostic system. Fig. 6 illustrates an overview diagram that directly maps the conceptual structure from Fig. 5, while Fig. 7 provides a detailed description of the internal blocks within the "AMFL model" component shown in Fig. 6.

The proposed AMFL model incorporates a key enhancement by enabling the handling of large-scale input datasets in matrix format, as illustrated in the "Data_DGA" block in Fig. 6. This capability allows the system to simultaneously process extensive sets of dissolved gas data and compute both the intermediate outputs of individual diagnostic methods and the final integrated diagnostic result in parallel. The generated outputs are automatically exported to MATLAB workspace variables and subsequently transferred to an Excel file via a callback function, thereby improving the system's practical applicability and user-friendliness for industrial implementation.

At the core of this framework is the fuzzy inference engine, which generates preliminary fault classifications based on optimized membership functions. These classifications are derived from preprocessed DGA input data, which are fed into individual FL blocks representing each diagnostic method. The output from these FL blocks is aligned with the sequential diagnostic logic outlined in the algorithm flowchart in Fig. 5, which is $D_1, D_2, D_3, D_4, D_5$.

These individual outputs are subsequently integrated into the model evaluation. In which a validation step is performed, implemented via MATLAB's "Compare" function blocks in Fig. 7, to filter out invalid diagnostic codes and ensure that only valid results are processed in the next stage. The validated outputs are then forwarded to compute the final decision, which is then input into the weight adaptation module. This module assigns and updates the contribution of each method based on their diagnostic performance utilizing Algorithm 1.

This weighted decision-making process not only improves the reliability and robustness of the system but also enhances its adaptability to different fault scenarios. The Simulink-based simulation thus provides a comprehensive representation of the entire diagnostic pipeline, demonstrating how data preprocessing, fuzzy inference, method integration, and decision-making are seamlessly coordinated within the proposed framework.

### *4.4. Experimental results and comparative analysis*

In this study, the training and evaluation processes were conducted using MATLAB R2024b on a standard laptop equipped with an AMD Ryzen 5 5500 U (2.1 GHz), Radeon Graphics, and 24 GB of RAM. For this configuration and a dataset of 600 DGA samples, the total training time required for weight optimization in the AMFL model was approximately 2 seconds. This demonstrates the model's computational efficiency and its suitability for real-time or embedded diagnostic systems.

The proposed system underwent validation using a comprehensive dataset comprising historical DGA records with confirmed fault classifications. To assess the accuracy of the proposed AMFL approach, a comparison was made between the diagnostic accuracy of traditional methods and that of the AMFL model. The simulation was performed in accordance with Subsection 3.3. After the simulation, weights and accuracy rates of methods after each iteration are presented in Table 7 with the AMFL model achieving an accuracy of 99.4%. The weights were adapted from their initial values, immediately adjusting after the first iteration to accommodate the training dataset of 600 data points. As iterations progressed, the adjustments to the weights gradually diminished, eventually reaching a point of stabilization where no further changes were necessary.

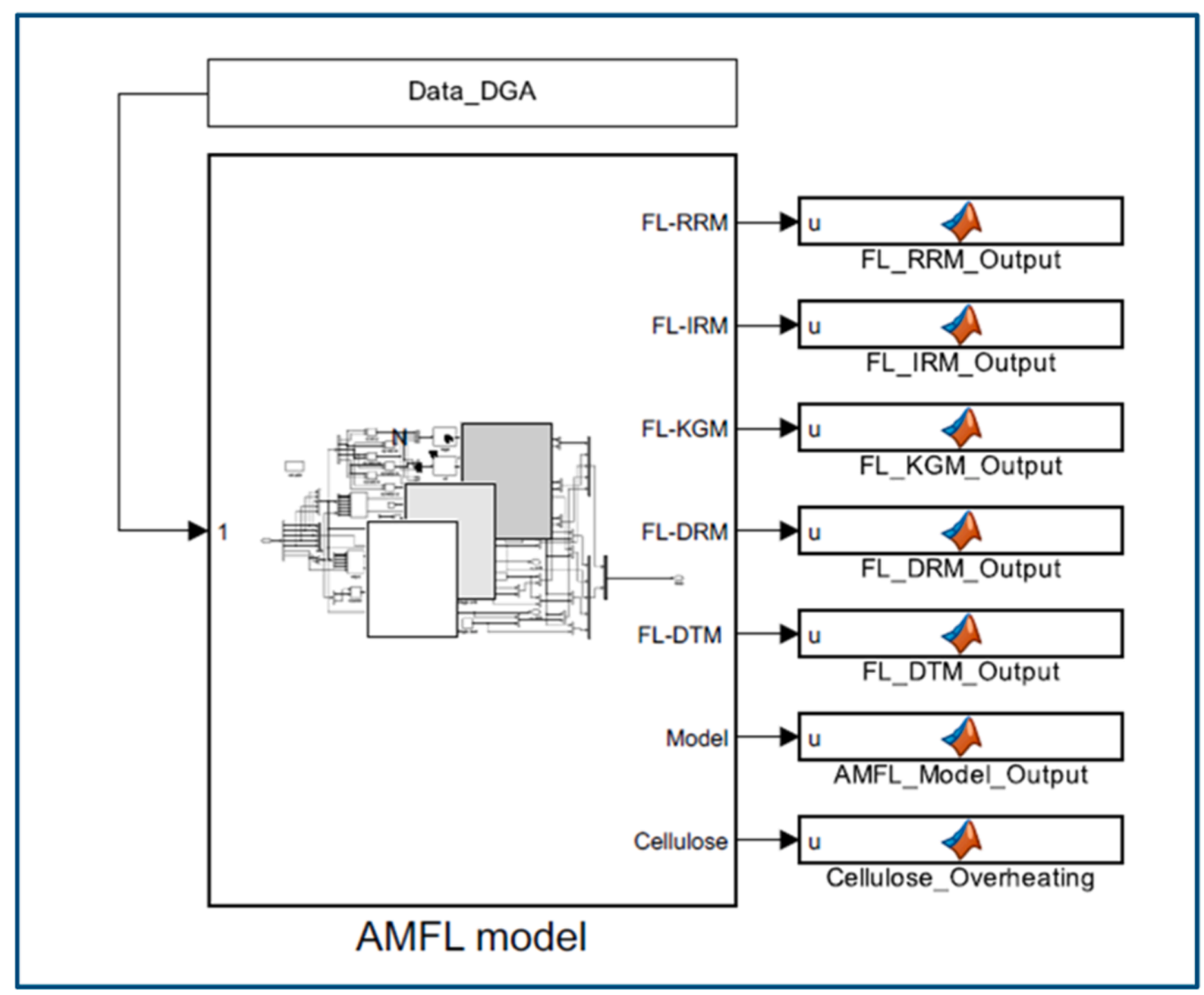

**Fig. 6.** Simulink block diagram of the AMFL model showing the DGA data input and individual fuzzy logic module outputs.

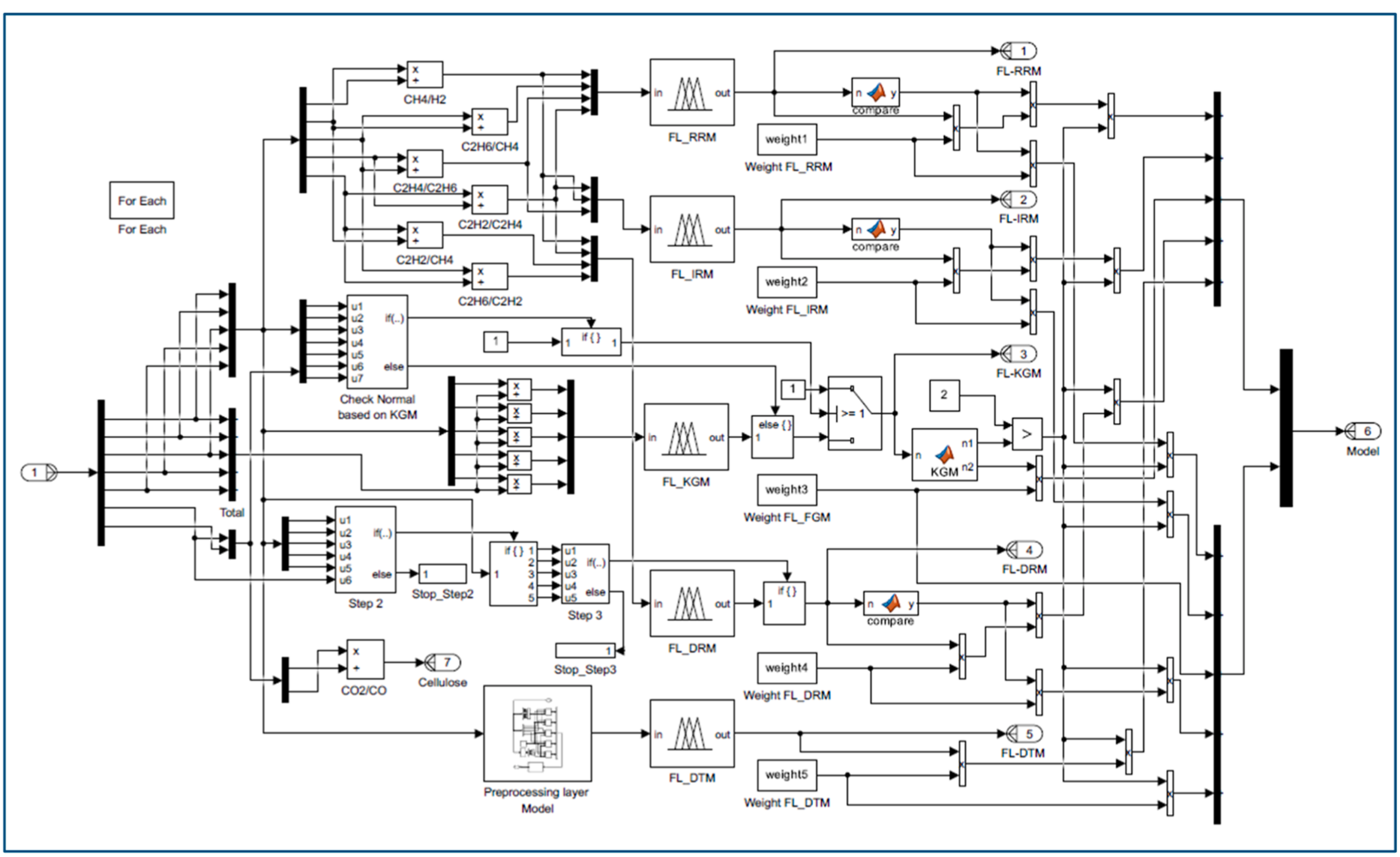

**Fig. 7.** Overall simulation of the AMFL model implemented in MATLAB/Simulink.

**Table 7**
Weights and accuracy rate of methods after each iteration.

| Weight | Iter. 1 | Iter. 2 | Iter. 3 | Iter. 4 | … | Iter. 13 | Iter. 14 | Iter. 15 | Iter. 16 | Iter. 17 | Iter. 18 | Iter. 19 | … | Iter. 49 | Iter. 50 |
|---|---|---|---|---|---|---|---|---|---|---|---|---|---|---|---|
| $W_1$ (RRM) | 0.1500 | 0.1427 | 0.1380 | 0.1348 | … | 0.1289 | 0.1288 | 0.1286 | 0.1285 | 0.1285 | 0.1285 | 0.1285 | … | 0.1285 | 0.1285 |
| $W_2$ (IRM) | 0.1551 | 0.1574 | 0.1591 | 0.1604 | … | 0.1634 | 0.1635 | 0.1635 | 0.1637 | 0.1637 | 0.1637 | 0.1637 | … | 0.1637 | 0.1637 |
| $W_3$ (KGM) | 0.2600 | 0.2672 | 0.2720 | 0.2752 | … | 0.2812 | 0.2813 | 0.2815 | 0.2816 | 0.2816 | 0.2816 | 0.2816 | … | 0.2816 | 0.2816 |
| $W_4$ (DRM) | 0.1710 | 0.1563 | 0.1464 | 0.1396 | … | 0.1260 | 0.1257 | 0.1254 | 0.1250 | 0.1250 | 0.1250 | 0.1250 | … | 0.1250 | 0.1250 |
| $W_5$ (DTM) | 0.2639 | 0.2763 | 0.2846 | 0.2900 | … | 0.3005 | 0.3007 | 0.3010 | 0.3012 | 0.3012 | 0.3012 | 0.3012 | … | 0.3012 | 0.3012 |
| **Accuracy**$(A)$ | **96.9%** | **96.9%** | **97.5%** | **97.5%** | **…** | **99.4%** | **99.4%** | **99.4%** | **99.4%** | **99.4%** | **99.4%** | **99.4%** | **…** | **99.4%** | **99.4%** |

The *MSE* achieved its lowest value of 0.075 at iteration 20, as depicted in Fig. 8. This outcome indicates that the weights were effectively optimized and had converged to their optimal values, as illustrated in Fig. 9. The results are presented in Table 8. It can be observed that the accuracy of traditional methods is relatively low, failing to demonstrate consistency and high reliability. In contrast, the enhanced FL-DGA model significantly improves diagnostic accuracy, with an improvement of approximately 10-20% for each method. Notably, the FL-RRM and FL-KGM achieve accuracies of 82.5% and 97.5%, respectively, which are substantially higher than their traditional counterparts, which only achieve approximately 37.8% and 57.8%. This marked improvement highlights the effectiveness and reliability of the FL-DGA model in fault diagnosis for power transformers. As shown, Fig. 10 illustrates the comparison of diagnostic accuracy between traditional methods and the FL-DGA model, further emphasizing the improvements achieved.

To further evaluate the accuracy and robustness of the proposed AMFL model, an assessment of its adaptability to various datasets was conducted. The existing traning dataset of 600 data points was divided into five distinct subsets, as shown in Table 9, each designed with unique fault distributions to simulate diverse operating conditions. Subset 1 predominantly contained data representing electrical faults, while subset 2 focused primarily on thermal faults. Subset 3 was characterized by a large proportion of PD cases, whereas subset 4 had an evenly distributed mix of all fault types. Finally, subset 5 consisted of 120 randomly selected data points to ensure variability.

The evaluation results, illustrated in Figs. 11 and 12, highlight that the weight mechanism significantly surpassed the fixed-weight model across all subsets. This improvement was particularly notable in subsets with imbalanced fault distributions, such as subsets 1, 2, and 3, where traditional models exhibited limited accuracy and struggled with consistent fault classification By dynamically recalibrating the weights based on the fault characteristics within each subset, the weight system provided tailored and precise outputs for every condition. For instance, in subsets 1 and 2, where faults were predominantly electrical or thermal, the model demonstrated superior performance by adjusting the weights of the respective diagnostic methods, thereby enhancing classification accuracy. Similarly, in subset 3, where an imbalance of PD cases posed significant challenges, the adaptive model maintained consistent accuracy by reallocating weights to mitigate misclassification.

These results emphasize a key advancement of the proposed FL model: its remarkable adaptability to datasets with varying fault distributions. Unlike the fixed-weight approach, which is inherently limited by static configurations, the adaptive weight mechanism ensures robust performance even in complex and heterogeneous fault scenarios. This innovation not only elevates diagnostic accuracy by optimizing the decision-making process but also demonstrates the model's practicality and reliability for real-world applications, where fault distributions are often unpredictable and non-uniform. The results underscore the superiority of the AMFL model in addressing the limitations of traditional methods, reinforcing its potential for effective transformer condition monitoring in diverse operational environments.

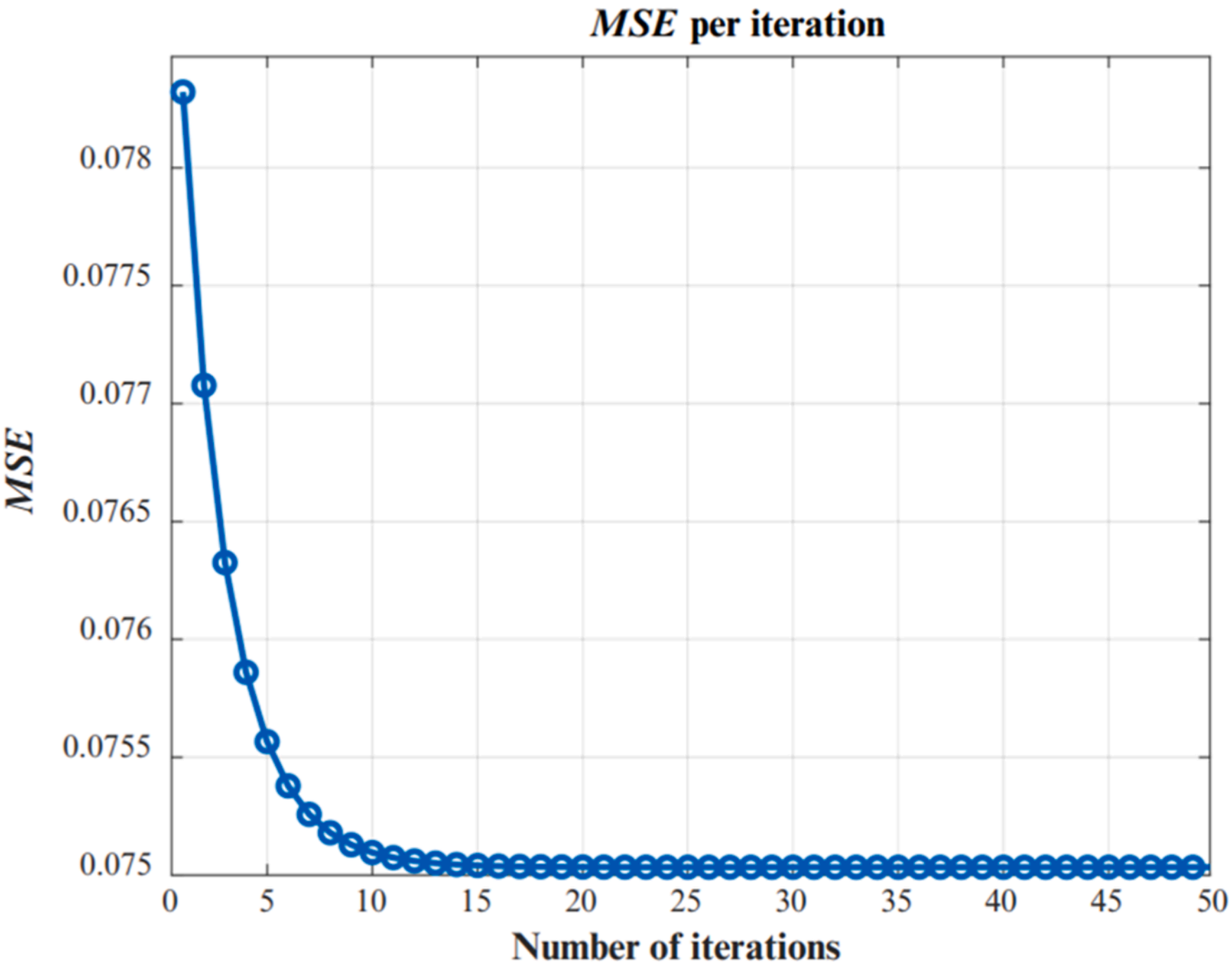

**Fig. 8.** MSE against 50 iterations during the training process of the AMFL model.

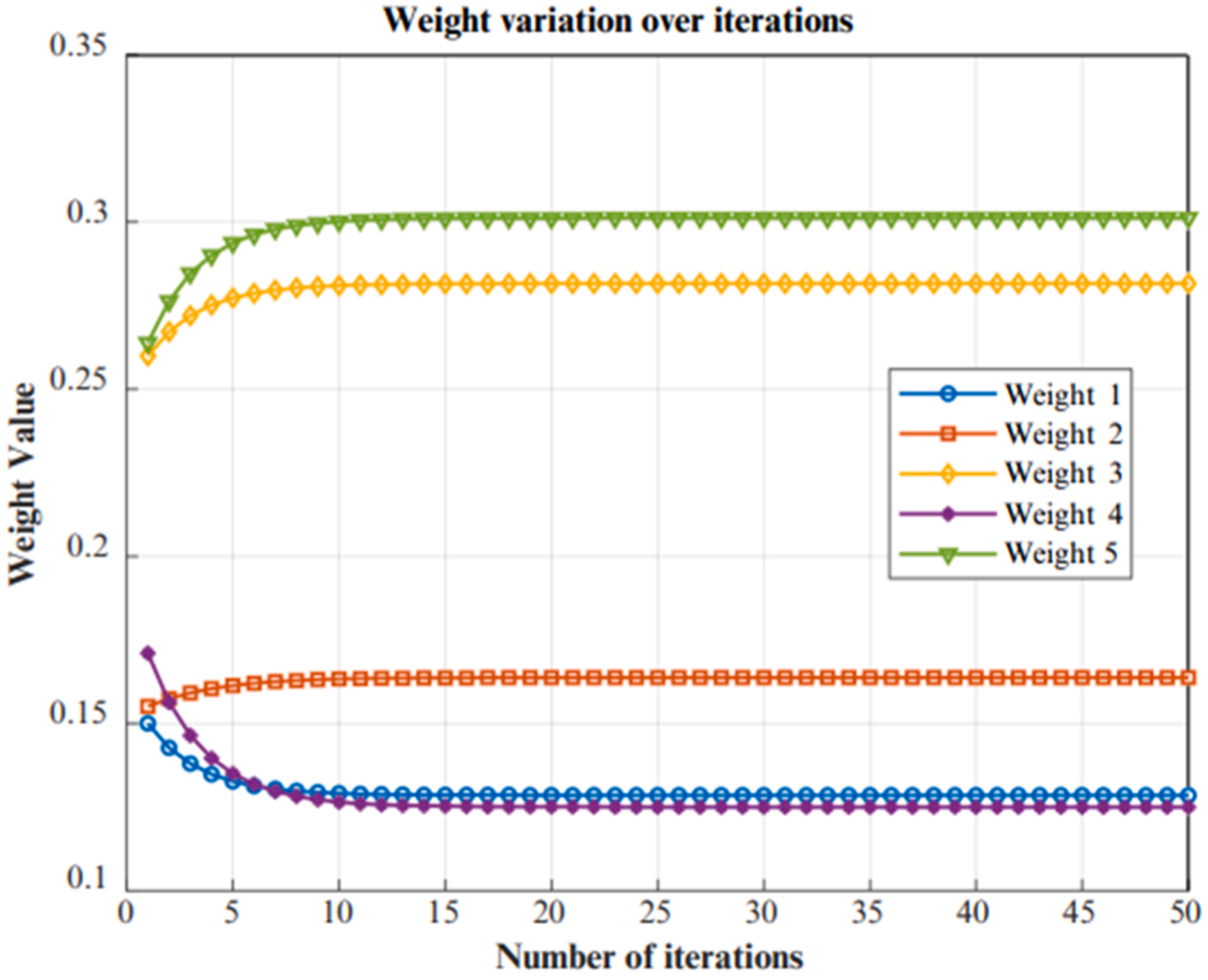

**Fig. 9.** Weight values adapted after each iteration.

**Table 8**
Accuracy comparison between traditional methods and their improved FL-DGA model.

| Method | DTM | | RRM | | IRM | | DRM | | KGM | | AMFL model |
|---|---|---|---|---|---|---|---|---|---|---|---|
| | Tra. | FL | Tra. | FL | Tra. | FL | Tra. | FL | Tra. | FL | |
| Total cases | 760 | 160 | 760 | 160 | 760 | 160 | 760 | 160 | 760 | 160 | 160 |
| Successful predicted case | 609 | 157 | 287 | 132 | 451 | 131 | 224 | 72 | 439 | 156 | 159 |
| **Accuracy** | 80.2% | 98.1% | 37.8% | 82.5% | 59.4% | 81.8% | 29.5% | 45.0% | 57.8% | 97.5% | 99.4% |

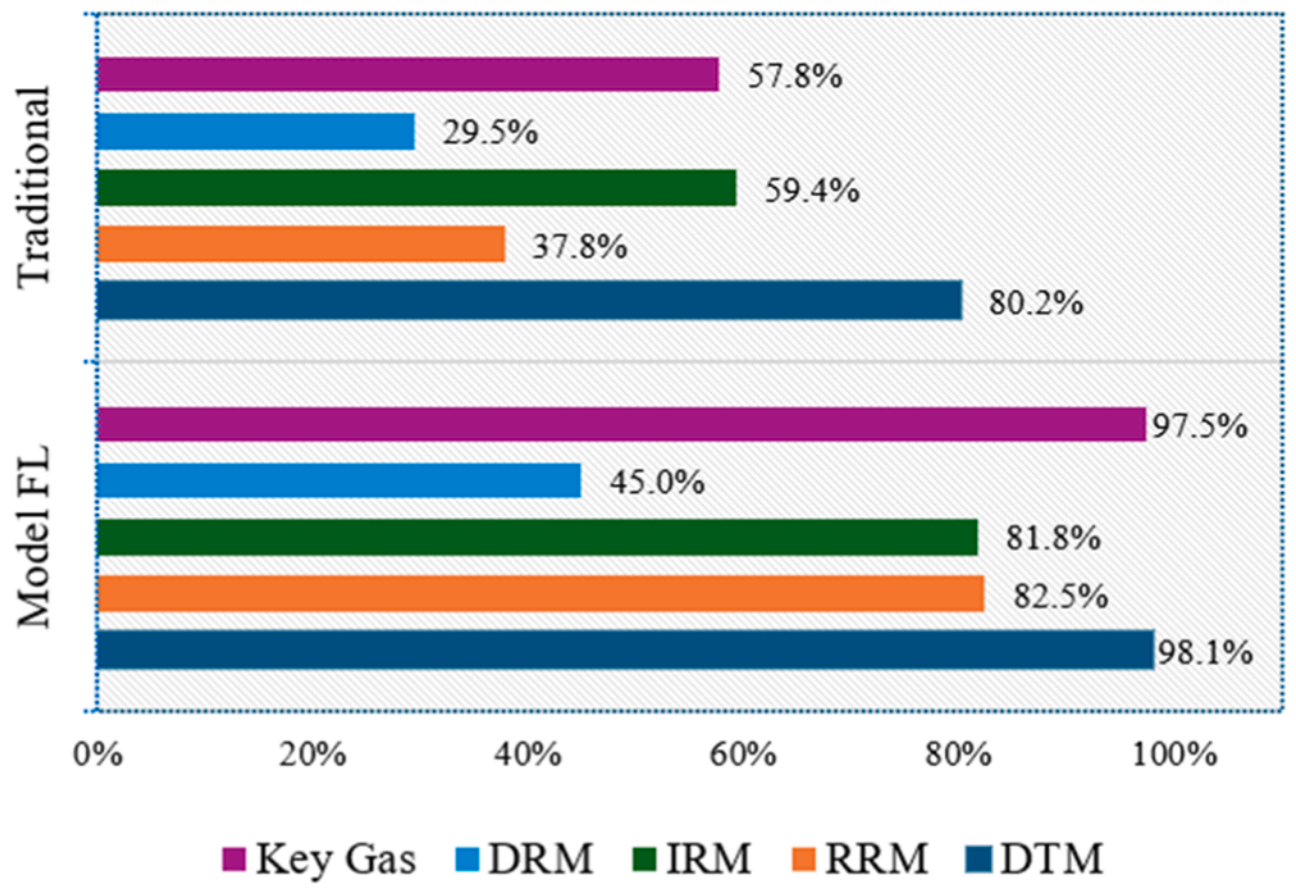

**Fig. 10.** Comparison results between traditional methods and improved FL-DGA model.

**Table 9**
Fault distribution of five subsets.

| Subset | Fault types | | | Total |
|---|---|---|---|---|
| | Thermal | Discharge | PD | |
| 1 | 27 | 88 | 5 | 120 |
| 2 | 85 | 26 | 9 | 120 |
| 3 | 38 | 33 | 49 | 120 |
| 4 | 56 | 57 | 7 | 120 |
| 5 | 67 | 47 | 6 | 120 |

To assess the robustness of the proposed AMFL model, experiments were conducted on eight transformer oil data points previously analyzed in [1]. The results of these experiments, presented in Table 10 and Table 11, demonstrate the significant improvements achieved by the proposed approach compared to traditional methods and models from previous studies.

Table 10 presents a comparative analysis between the diagnostic results reported in [1] and those obtained using the proposed AMFL model. For samples 1 and 2, which represent normal operating conditions with low concentrations of key gases, both approaches successfully classified them as normal (F4), demonstrating consistency with the actual fault conditions. In the case of samples 3 and 4, the multi-FL-DGA model from [1] identified F1 as the likely fault type; however, some variations were observed across its individual FL components—such as differing outputs from FL-KGM, FL-DRM, and FL-IRM. In contrast, the proposed AMFL model, benefiting from membership functions and weighting, provided consistent and accurate classification of both samples as thermal faults (F1). The corresponding diagnostic scores, 2.98 and 3.01 respectively (as shown in Table 11), reflect a high level of precision in fault severity estimation.

Samples 5 and 6, previously labeled as PD faults in [1], were further examined under the AMFL framework. In particular, sample 5 exhibited dominant concentrations of $CH_4$ and $C_2H_4$—typically associated with thermal faults—and was accordingly reclassified as F1 with a score of 2.97. Meanwhile, sample 6 presented a more complex condition, with high levels of both $C_2H_2$ (indicating discharges) and $CH_4/C_2H_4$ (suggesting thermal involvement). Leveraging its integrated multi-method architecture, the AMFL model effectively recognized this as a combined thermal-electrical fault (F6), with a score of 3.76—highlighting its ability to capture nuanced fault profiles.

For samples 7 and 8, both models identified discharge-related issues, with the AMFL model producing consistent F3 classifications and corresponding fault rankings of 7.01 and 7.00. These results collectively illustrate the diagnostic improvements achieved by the proposed model, particularly in handling complex or overlapping fault conditions, while also affirming the valuable contributions of prior approaches that laid the groundwork for these advancements.

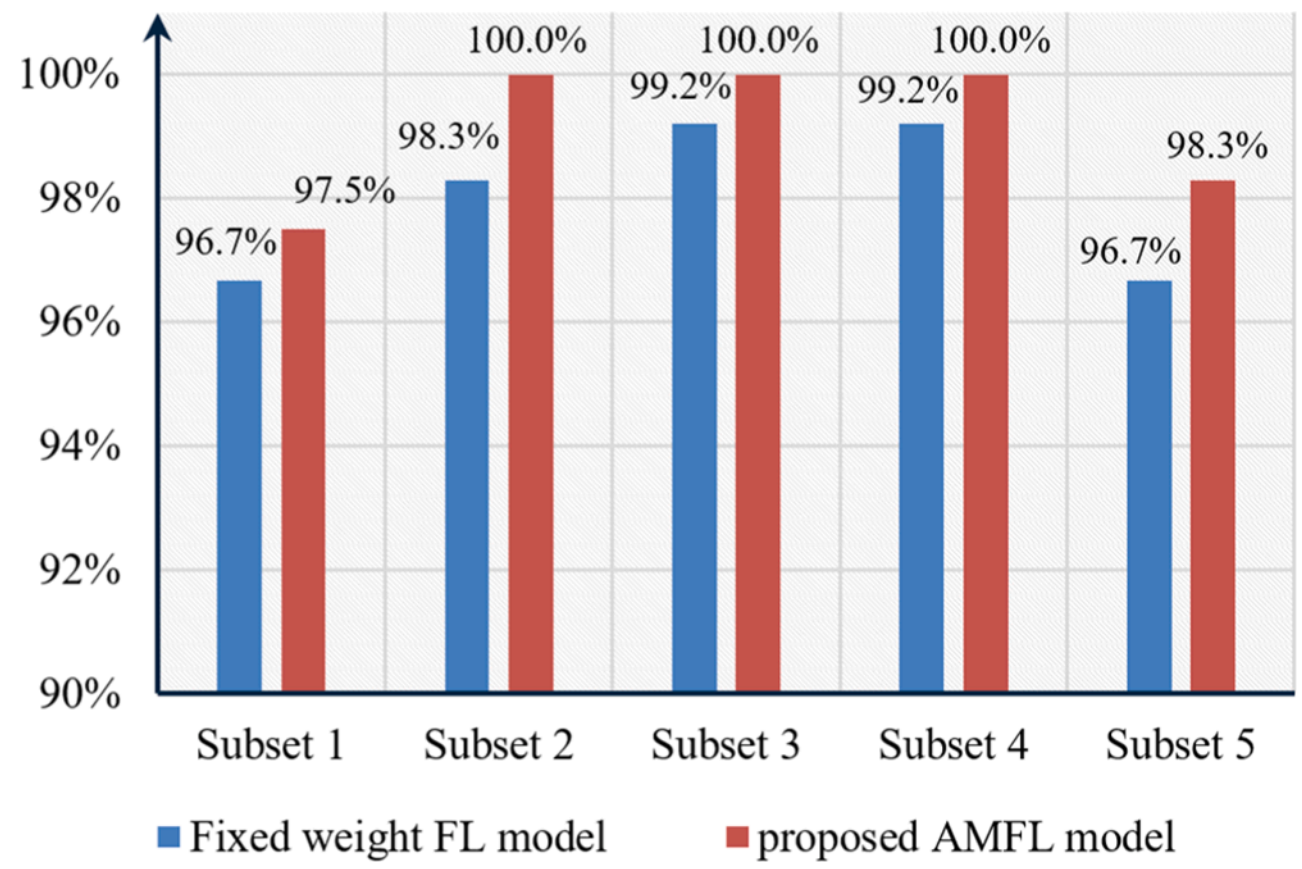

**Fig. 12.** Comparison results of different subsets between fixed weight multi-FL-DGA model and proposed AMFL model.

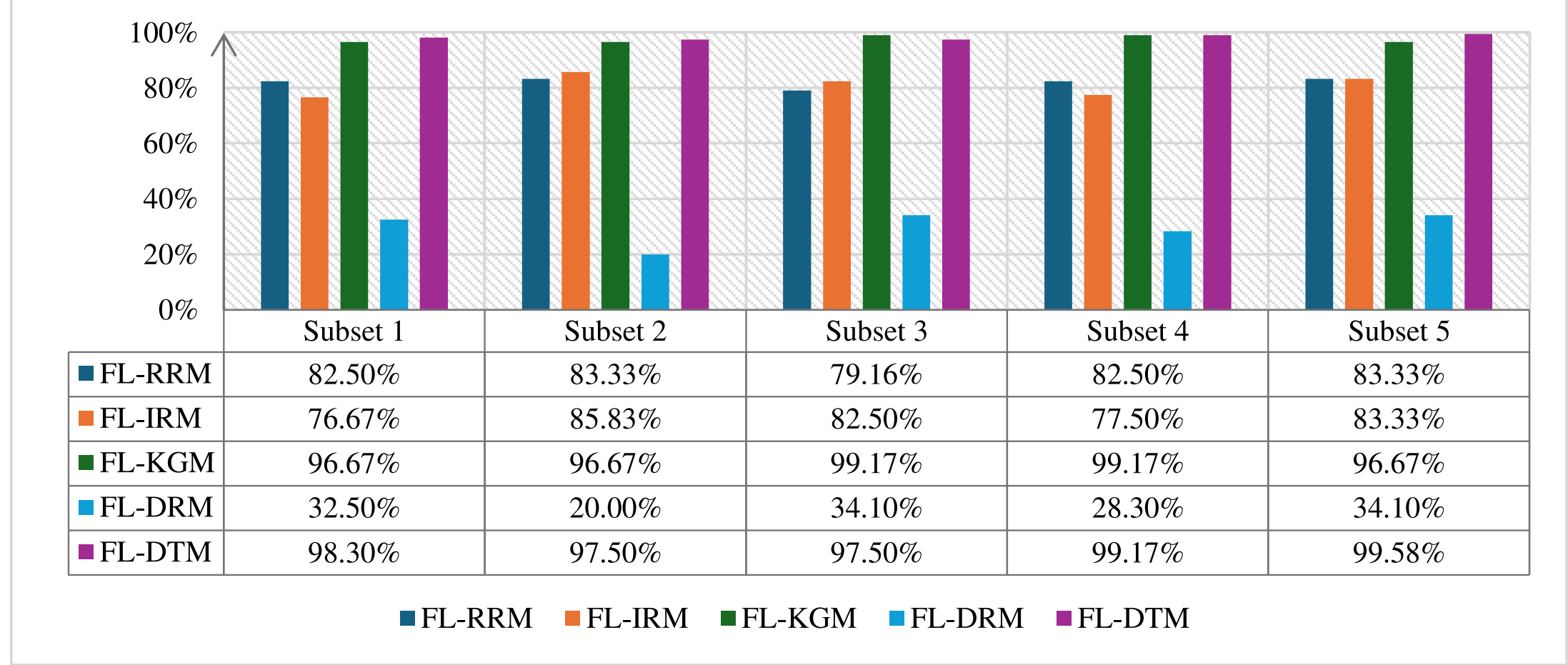

| | Subset 1 | Subset 2 | Subset 3 | Subset 4 | Subset 5 |
|---|---|---|---|---|---|
| FL-RRM | 82.50% | 83.33% | 79.16% | 82.50% | 83.33% |
| FL-IRM | 76.67% | 85.83% | 82.50% | 77.50% | 83.33% |
| FL-KGM | 96.67% | 96.67% | 99.17% | 99.17% | 96.67% |
| FL-DRM | 32.50% | 20.00% | 34.10% | 28.30% | 34.10% |
| FL-DTM | 98.30% | 97.50% | 97.50% | 99.17% | 99.58% |

**Fig. 11.** Accuracy achieved at different subsets at different FL-DGA methods.

**Table 10**
Diagnostic results from the methods in the study [1].

| Sample | Gases in ppm | | | | | | | FL | | | | | Actual fault | Model output |
|---|---|---|---|---|---|---|---|---|---|---|---|---|---|---|
| | $H_2$ | $CH_4$ | $C_2H_2$ | $C_2H_4$ | $C_2H_6$ | CO | $CO_2$ | KGM | DTM | RRM | DRM | IRM | | |
| 1 | 2 | 7 | 0 | 0 | 0 | 0 | 132 | F4 | F2 | F5 | F1 | F5 | Normal | 1 (F4) |
| 2 | 47 | 12 | 0 | 8 | 0 | 115 | 1113 | F4 | F1 | F5 | F3 | F5 | Normal | 1.5 (F4) |
| 3 | 124 | 166 | 0 | 59 | 87 | 530 | 3750 | F2 | F1 | F1 | F1 | F1 | Heating | 3.426 (F1) |
| 4 | 103 | 74 | 0 | 9 | 80 | 754 | 2605 | F1 | F1 | F1 | F3 | F4 | Heating | 3.3 (F1) |
| 5 | 231 | 3997 | 0 | 5584 | 1726 | 0 | 2194 | F2 | F2 | F5 | F1 | F1 | PD | 4.4 (F2) |
| 6 | 384 | 388 | 33 | 110 | 55 | 173 | 932 | F2 | F2 | F5 | F1 | F3 | PD | 5.5 (F2) |
| 7 | 441 | 207 | 261 | 224 | 43 | 161 | 1,123 | F2 | F3 | F3 | F3 | F3 | Arcing | 6.5 (F3) |
| 8 | 217 | 286 | 884 | 458 | 14 | 176 | 1544 | F2 | F3 | F5 | F1 | F3 | Arcing | 7.5 (F3) |

**Table 11**
Improved diagnostic results from the proposed method.

| Sample | FL | | | | | | Traditional method | | | | | Actual fault |
|---|---|---|---|---|---|---|---|---|---|---|---|---|
| | KGM | DTM | RRM | DRM | IRM | MODEL | KGM | DTM | RRM | DRM | IRM | |
| 1 | F4 | F1 | F5 | F5 | F5 | 1.00 (F4) | N | PD | UD | UD | UD | Normal |
| 2 | F4 | F2 | F5 | F5 | F1 | 1.00 (F4) | N | T | UD | UD | UD | Normal |
| 3 | F1 | F1 | F1 | F5 | F1 | 2.98 (F1) | UD | T | T | UD | T | Heating |
| 4 | F1 | F1 | F1 | F5 | F4 | 3.01 (F1) | UD | T | T | UD | N | Heating |
| 5 | F1 | F1 | F5 | F1 | F1 | 2.97 (F1) | T | T | UD | T | T | Heating |
| 6 | F1 | F1 | F1 | F1 | F5 | 4.09 (F6) | UD | DT | T | T | UD | DT |
| 7 | F3 | F3 | F3 | F3 | F3 | 7.01 (F3) | D | D | D | D | D | Arcing |
| 8 | F3 | F3 | F3 | F5 | F5 | 7.00 (F3) | D | D | UD | UD | UD | Arcing |

The advancements realized by the AMFL model are attributed to its capacity to integrate multiple diagnostic parameters, facilitating a more nuanced and precise fault classification compared to the rigid single-ratio thresholds employed in [1]. By addressing the limitations inherent in previous methodologies, this model provides reliable diagnostic outcomes for both simple and complex fault conditions, as demonstrated by the improved classifications for samples 3, 5, and 6. This validation highlights the efficacy of the AMFL model in enhancing transformer fault diagnosis.

To summarize the findings of the proposed AMFL model, a comparative analysis was conducted by evaluating the AMFL model using the DGA dataset reported in each corresponding study as input. Specifically, for each referenced study in Table 12, the DGA dataset is used as the input for the AMFL model to generate the corresponding accuracy value. This ensures a consistent evaluation basis and allows for direct comparison of diagnostic performance.

The comparative results show that the AMFL model consistently outperforms several state-of-the-art techniques across multiple dataset sizes. For example, compared to CNN (98.50%) [25], Random Forest (96.09%) [50], and Hybrid ANFIS + DST (92.50%) [18], the proposed model achieved notably higher accuracies of 99.51%, 99.17%, and 99.25% respectively when tested on the same dataset samples used in those studies.

In cases where the data were incomplete or unavailable (for example, [15,20,23]), the diagnostic accuracy of the proposed model could not be evaluated and is indicated as "–". Nonetheless, for all references with accessible datasets, the proposed AMFL model achieved an accuracy of over 99 %, highlighting its superior adaptability and robustness in transformer fault diagnosis.

To further substantiate the effectiveness and robustness of the proposed approach, a selection of samples from the evaluation process is presented in Table 13. These samples illustrate the alignment between the diagnostic outputs and the actual fault conditions, thereby demonstrating the precision and quality of the AMFL model. The results indicate that the method not only maintains high accuracy across diverse datasets but also consistently produces outputs that closely correspond to the true fault types. This alignment underscores the robustness and practical value of the model, further validating its capacity to address the inherent challenges of transformer fault diagnostics with a high degree of confidence.

**Table 12**
Performance comparison of DGA models reported in the literature.

| Reference | Model | Accuracy (%) | Number of data points | Accuracy of the proposed AMFL model (%) | Remarks on the data |
|---|---|---|---|---|---|
| [15] | ANN | 87.80% | | – | *No data* |
| [18] | Hybrid ANFIS + DST | 92.50% | 134 | 99.25% | |
| [20] | Hybrid SVM+ SVFB | 97.35% | 419 | – | *Data not available* |
| [35] | Hybrid DPM + FL +SVM | 97.60% | 127 | – | *Data not available* |
| [39] | FLTFDS | 82.12% | 14 | 100 % | |
| [43] | FIS | 97.40% | 117 | 99.15% | |
| [23] | Hybrid ENN-SVM | 88.00% | | – | *No data* |
| [24] | Hybrid k-MCA-GA | 98.29% | 117 | 99.15% | |
| [25] | CNN | 98.50% | 206 | 99.51% | |
| [27] | TLBO | 88.86 % | 89 | 100% | |
| [50] | Random Forest | 96.09% | 121 | 99.17% | |

To provide a more complete picture of the model's behavior under edge conditions, it is worth noting that when the AMFL model was applied to 160 DGA samples from the testing dataset, only one instance, the 21$^{st}$ sample, represented a typical edge condition in which the model yielded a misclassification due to overlapping fault characteristics. Among the five fuzzy diagnostic modules, three returned UD outputs, while FL-KGM and FL-DTM produced diagnoses of PD and DT, respectively. Given the narrow interval weights associated with the DT category, the weighted aggregation favored PD. However, further inspection reveals that the sample exhibits hybrid fault indicators: elevated $H_2$ (PD), increased $C_2H_4$ (thermal), and the presence of $C_2H_2$ (electrical). Although the final classification differs from the ground truth, the model's decision remains explainable. This case highlights the advantage of fuzzy-based multi-perspective reasoning in capturing real-world diagnostic complexity more effectively than rigid binary classification.

**Table 13**
Examples of typical diagnostic outputs from the proposed AMFL model.

| Sample | Gas Input (ppm) | | | | | | | FL | | | | | | Traditional Method | | | | | Actual fault |
|---|---|---|---|---|---|---|---|---|---|---|---|---|---|---|---|---|---|---|---|
| | $H_2$ | $CH_4$ | C2H2 | C2H4 | C2H6 | CO | $CO_2$ | KGM | DTM | RRM | DRM | IRM | Model output | KGM | DTM | RRM | DRM | IRM | |
| 1 | 37800 | 1740 | 249 | 8 | 8 | 56 | 197 | F2 | F2 | F2 | F5 | F2 | 4.9 (F2) | PD | PD | PD | UD | PD | PD |
| 2 | 8266 | 1061 | 22 | 0.01 | 0.01 | 107 | 498 | F2 | F2 | F3 | F5 | F5 | 5.3 (F2) | PD | PD | D | UD | UD | PD |
| 3 | 9340 | 995 | 60 | 6 | 7 | 60 | 620 | F2 | F2 | F3 | F2 | F2 | 5.1 (F2) | PD | PD | D | PD | UD | PD |
| 4 | 33046 | 619 | 58 | 2 | 0 | 51 | 1 | F2 | F2 | F2 | F5 | F2 | 4.9 (F2) | PD | PD | PD | UD | PD | PD |
| 5 | 40280 | 1069 | 1060 | 1 | 1 | 1 | 0.01 | F2 | F2 | F2 | F5 | F2 | 4.8 (F2) | PD | PD | PD | UD | PD | PD |
| 6 | 2177 | 1049 | 207 | 440 | 705 | 4571 | 3923 | F3 | F3 | F3 | F3 | F3 | 7.0 (F3) | PD | D | D | D | D | Arcing |
| 7 | 3090 | 5020 | 323 | 3800 | 2540 | 270 | 400 | F3 | F3 | F3 | F5 | F5 | 7.0 (F3) | UD | D | D | UD | UD | Arcing |
| 8 | 110 | 62 | 90 | 140 | 250 | 680 | 6470 | F3 | F3 | F3 | F3 | F3 | 7.0 (F3) | D | D | UD | D | D | Arcing |
| 9 | 5000 | 1200 | 83 | 1000 | 1100 | 140 | 265 | F3 | F3 | F3 | F3 | F3 | 7.0 (F3) | PD | D | D | D | D | Arcing |
| 10 | 150 | 130 | 9 | 55 | 30 | 120 | 200 | F3 | F6 | F3 | F5 | F3 | 6.5 (F3) | PD | D | D | UD | D | Arcing |
| 11 | 1860 | 4980 | 1600 | 10700 | 1600 | 158 | 1300 | F1 | F1 | F1 | F3 | F5 | 3.8 (F1) | T | T | T | UD | UD | Heating |
| 12 | 400 | 940 | 24 | 820 | 24 | 390 | 1700 | F1 | F1 | F1 | F1 | F1 | 2.9 (F1) | UD | T | T | T | T | Heating |
| 13 | 290 | 1260 | 8 | 820 | 8 | 228 | 826 | F1 | F1 | F1 | F5 | F1 | 2.9 (F1) | UD | T | UD | UD | T | Heating |
| 14 | 6 | 2990 | 67 | 26076 | 67 | 6 | 26 | F1 | F1 | F1 | F1 | F1 | 4.1 (F1) | T | T | UD | T | T | Heating |
| 15 | 1 | 8 | 6 | 100 | 6 | 300 | 5130 | F1 | F1 | F1 | F5 | F1 | 2.9 (F1) | T | T | UD | UD | T | Heating |
| 16 | 150 | 22 | 9 | 60 | 11 | 21 | 140 | F1 | F1 | F5 | F5 | F3 | 4.1 (F6) | T | T | UD | UD | D | DT |
| 17 | 105 | 23 | 13 | 4 | 3 | 15 | 107 | F2 | F6 | F3 | F5 | F5 | 5.7 (F6) | PD | DT | D | UD | UD | DT |
| 18 | 18.2 | 22 | 6.6 | 46.9 | 4 | 35 | 420 | F3 | F1 | F1 | F5 | F1 | 4.2 (F6) | D | T | T | UD | T | DT |
| 19 | 0 | 1 | 0 | 4 | 0 | 10 | 200 | F4 | F1 | F2 | F5 | F1 | 1.0 (F4) | N | UD | T | UD | T | Normal |
| 20 | 2 | 2 | 0 | 3 | 0 | 7 | 335 | F4 | F1 | F2 | F5 | F1 | 1.0 (F4) | N | T | T | UD | UD | Normal |
| **21** | **301** | **5.8** | **4.7** | **9.7** | **5** | **0.0001** | **0.0001** | **F2** | **F6** | **F5** | **F5** | **F5** | **5.4 (F2)** | **UD** | **DT** | **UD** | **UD** | **UD** | **Arcing** |

## 4. Conclusion

This article proposes an AMFL model for transformer fault diagnosis using a DGA, with the aim of enhancing diagnostic accuracy and practicality. The proposed method integrates the strengths of five conventional DGA interpretation methods—IRM, RRM, DRM, DTM, and KGM—by unifying them under a single fuzzy logic umbrella. Some of these prominent advancements include a dynamic weight optimization system that prefers more reliable diagnostic sources, reordered fuzzy membership functions with overlapping ranges to handle boundary and uncertain situations, and an enhanced decision space to support sparse and hybrid fault types such as DT faults. Such enhancements allow the model to identify diagnostic patterns that are more complex than are possible with traditional or static fuzzy approaches. There is large-scale experimental evidence that the AMFL model has an outstanding diagnostic accuracy of over 99% and outperforms single DGA methods and non-adaptive fuzzy logic-based models. The system is stable by continually dealing with overlapping or uncertain fault regions and arriving at the right conclusion even when some of the diagnostic modules yield "undetermined" results, only being compromised in the rare case when all the approaches fail simultaneously. Its ability to process unbalanced or incomplete datasets directly, and its implementation within MATLAB/Simulink, are evidence of the flexibility and practicality of the model for real-world use in real transformer monitoring systems. Its improvements allow the AMFL model to significantly enhance transformer fault diagnosis accuracy and contribute to more reliable condition monitoring.

Building upon the adaptive optimization mechanism introduced in the AMFL model, future research could explore the integration of Quantum Machine Learning (QML) to further enhance transformer fault diagnosis. By encoding classical DGA data into a high-dimensional Hilbert space via quantum feature maps, QML allows for more expressive and separable representations of complex fault patterns. A promising direction involves incorporating a Variational Quantum Classifier (VQC) into the AMFL framework—specifically at the stage of final fault classification based on the aggregated fuzzy outputs from multiple logic modules. This quantum-enhanced classifier can improve the accuracy of decision-making under overlapping or ambiguous conditions. Additionally, VQC can be explored as a component in the adaptive weight adjustment mechanism, helping optimize the contribution of each module with faster convergence and greater resilience to noise. Extending AMFL with VQC would reinforce its scalability, adaptability, and diagnostic accuracy across various transformer insulation systems and operational environments.

## CRediT authorship contribution statement

**Kim-Anh Nguyen:** Writing – review & editing, Supervision, Project administration, Methodology, Funding acquisition, Conceptualization. **Huy Hoang Le:** Writing – original draft, Validation, Software, Investigation, Formal analysis. **Ba Tu Phung:** Writing – original draft, Validation, Software, Resources, Data curation.

## Declaration of competing interest

The authors declare that they have no known competing financial interests or personal relationships that could have appeared to influence the work reported in this paper.

**Funding sources**

This work was supported by The University of Danang–University of Science and Technology, code number of Project: T2024-02-39.

## Appendix A

Fig. A1, Fig. A2, Fig. A3, Fig. A4

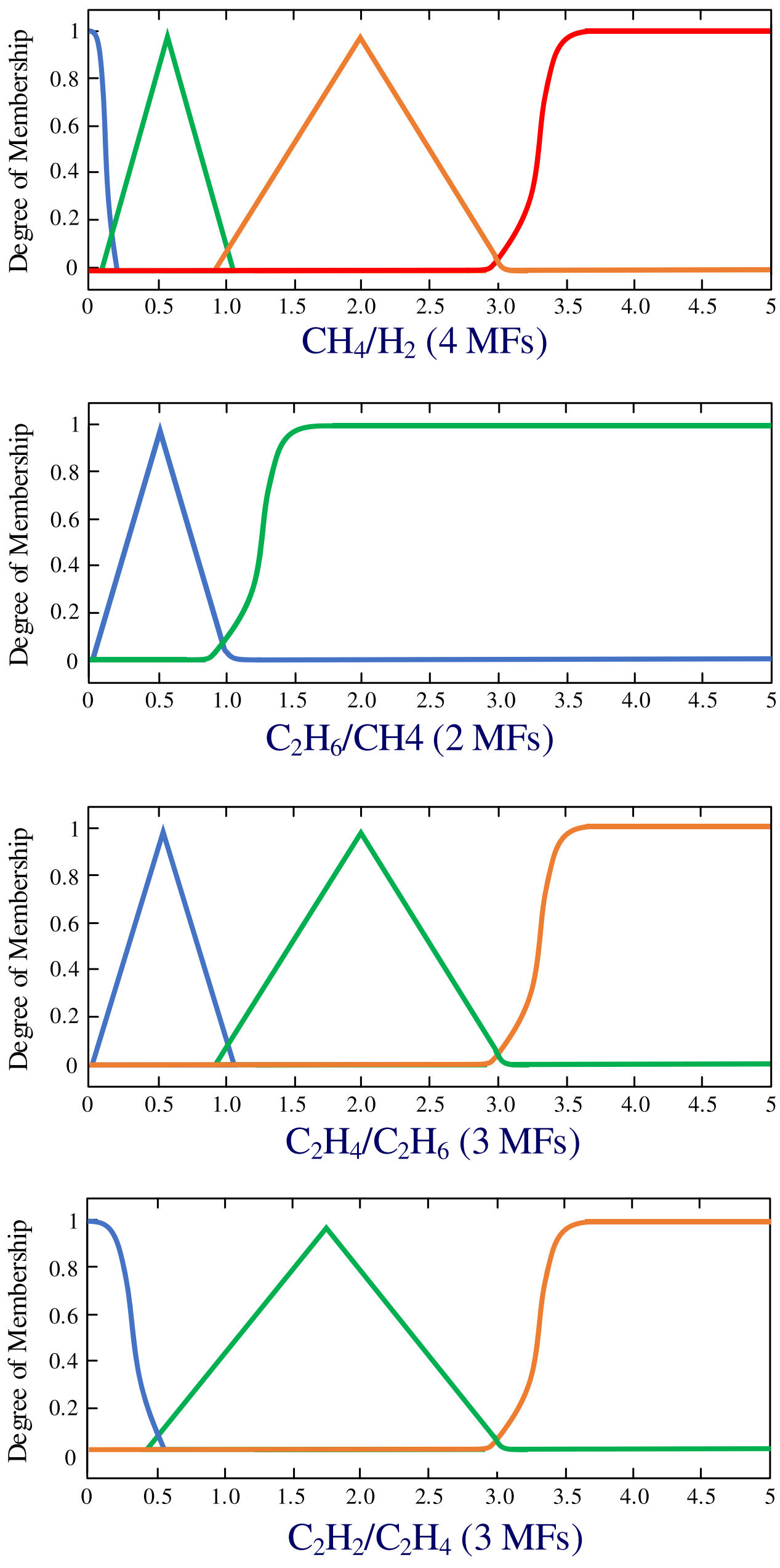

**Fig. A1.** Redesigned membership functions for the FL-RRM.

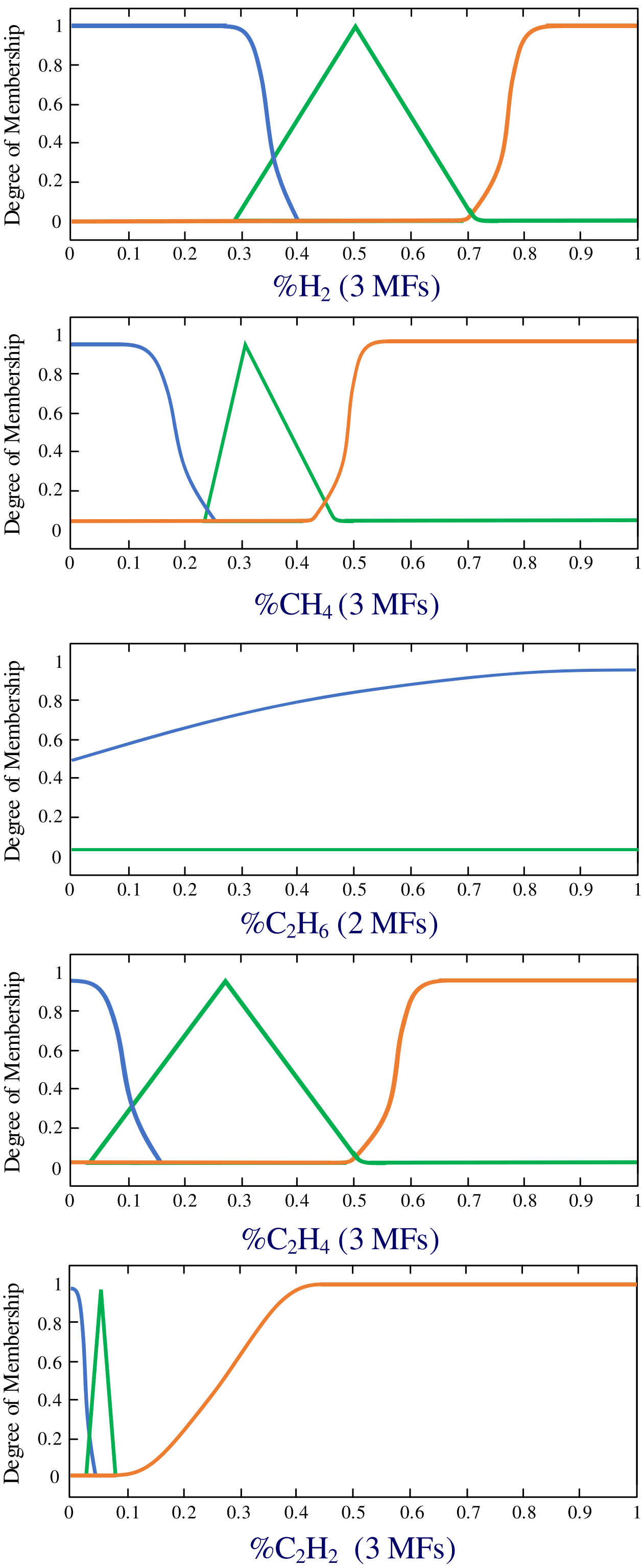

**Fig. A2.** Redesigned membership functions for the FL-KGM.

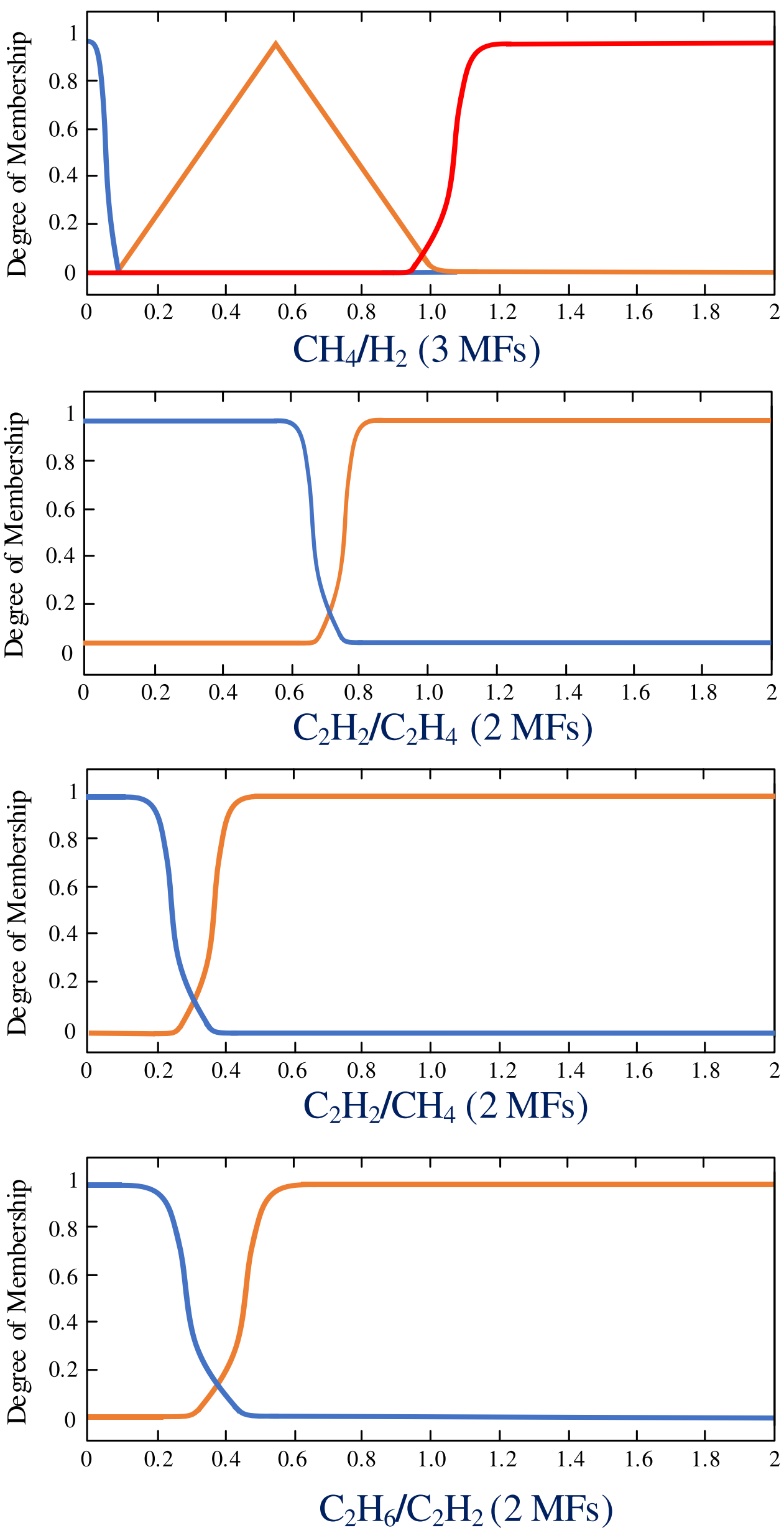

**Fig. A3.** Redesigned membership functions for the FL-DRM.

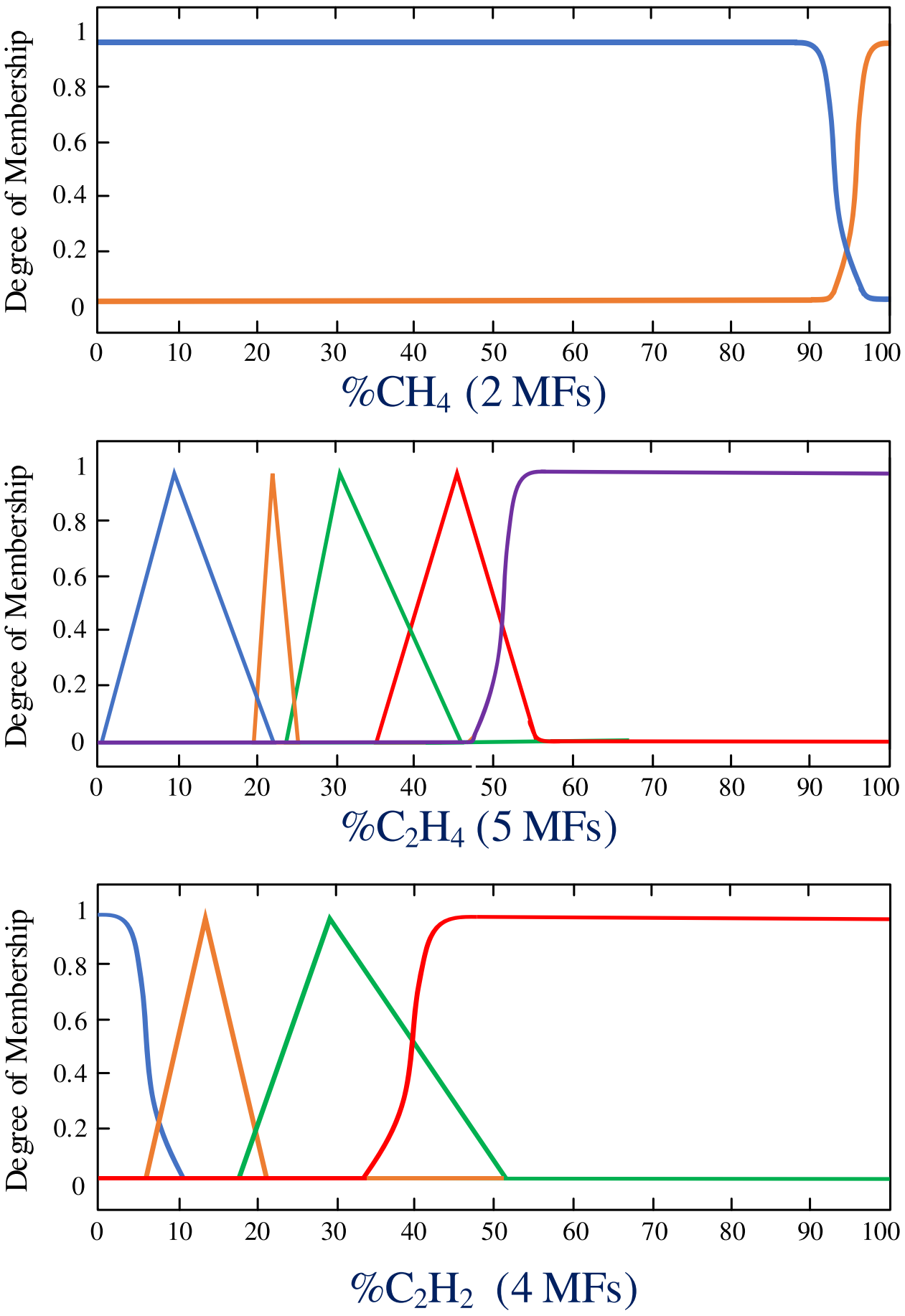

**Fig. A4.** Redesigned membership functions for the FL-DTM.

## Appendix B. An illustrative example of how Algorithm 1 operates in the k-th iteration

To provide better clarity on the optimization mechanism behind Algorithm 1, we present a detailed numerical example illustrating a complete iteration (e.g., $k = 1$) of a DGA dataset of 6 samples (noted that in our practical simulation, we used a training dataset of 600 DGA samples). This example demonstrates how the adaptive weight adjustment is performed, starting from initialization to weight update, with all key steps highlighted.

**Step 1** provide the initial parameters, where: $W_{init} = [0.5 \quad 0.517 \quad 0.867 \quad 0.57 \quad 0.88]$ present the initial weights of five FL-DGA modules as following: FL-RRM, FL-IRM, FL-KGM, FL-DRM, and FL-DTM; $W = [W_1, W_2, W_3, W_4, W_5]$ present the variance parameters of five FL-DGA's weights that can be update in the subsequent iterations; The decision output $D$ of the five FL-DGA modules is organized in the following sequence of five columns: FL-RRM, FL-IRM, FL-KGM, FL-DRM, and FL-DTM. And *actual_fault* to $D_{real}$ of each sample is represented in individual rows, matching the decision output $D$:

$$D = \begin{bmatrix} 3.3 & 5.2 & 2.9 & 3.0 & 3.1 \\ 7.5 & 7.0 & 6.9 & 7.2 & 6.8 \\ 3.1 & 3.0 & 3.5 & 2.8 & 2.7 \\ 4.7 & 5.0 & 5.3 & 5.2 & 4.5 \\ 5.9 & 5.3 & 4.8 & 3.9 & 4.7 \\ 7.2 & 7.9 & 6.8 & 6.1 & 6.5 \end{bmatrix}; \quad actual_fault = \begin{bmatrix} \text{T} \\ \text{D} \\ \text{T} \\ \text{PD} \\ \text{PD} \\ \text{D} \end{bmatrix} \Rightarrow D_{real} = \begin{bmatrix} 3.0 \\ 7.0 \\ 3.0 \\ 5.0 \\ 5.0 \\ 7.0 \end{bmatrix};$$

Other initial parameters are defined as:

$threshold = 10^{-5}$, $max_iteration = 100$, $\eta = 0.1$, $\delta = 2$;

**Step 2** is intended to set the initial value of variable $W$ in the first iteration (note that $W$ may change in subsequent iterations depending on the dataset):$W = W_{init} = [0.5 \quad 0.517 \quad 0.867 \quad 0.57 \quad 0.88]$;

**Step 3** starts the iteration process with $k = 1$ (i.e., the first iteration).

**Step 4** computes the value of $D_{combined,1}$, where:

$$\text{sum}(W) = 3.334 \quad \text{and} \quad D \times W^T = \begin{bmatrix} 3.3 & 5.2 & 2.9 & 3.0 & 3.1 \\ 7.5 & 7.0 & 6.9 & 7.2 & 6.8 \\ 3.1 & 3.0 & 3.5 & 2.8 & 2.7 \\ 4.7 & 5.0 & 5.3 & 5.2 & 4.5 \\ 5.9 & 5.3 & 4.8 & 3.9 & 4.7 \\ 7.2 & 7.9 & 6.8 & 6.1 & 6.5 \end{bmatrix} \times \begin{bmatrix} 0.500 \\ 0.517 \\ 0.867 \\ 0.570 \\ 0.880 \end{bmatrix} \approx \begin{bmatrix} 11.2907 \\ 22.9617 \\ 10.1136 \\ 17.0506 \\ 16.4766 \\ 22.3736 \end{bmatrix}$$

$$\Rightarrow \quad D_{combined,1} = \frac{D \times W^T}{\text{sum}(W)} \approx \begin{bmatrix} 3.3876 \\ 6.8916 \\ 3.0333 \\ 5.1161 \\ 4.9423 \\ 6.7111 \end{bmatrix}$$

**Step 5** computes the value of $Error_1$,

$$Error_1 = \left(D_{combined,1} - D_{real}\right)^2 = \left( \begin{bmatrix} 3.3876 \\ 6.8916 \\ 3.0333 \\ 5.1161 \\ 4.9423 \\ 6.7111 \end{bmatrix} - \begin{bmatrix} 3.000 \\ 7.000 \\ 3.000 \\ 5.000 \\ 5.000 \\ 7.000 \end{bmatrix} \right)^2 = \begin{bmatrix} 0.1502 \\ 0.0118 \\ 0.0011 \\ 0.0135 \\ 0.0033 \\ 0.0835 \end{bmatrix}$$

**Step 6** determines the value of *gradient* (i.e., approximated value), where:

$$D = \begin{bmatrix} 3.3 & 5.2 & 2.9 & 3.0 & 3.1 \\ 7.5 & 7.0 & 6.9 & 7.2 & 6.8 \\ 3.1 & 3.0 & 3.5 & 2.8 & 2.7 \\ 4.7 & 5.0 & 5.3 & 5.2 & 4.5 \\ 5.9 & 5.3 & 4.8 & 3.9 & 4.7 \\ 7.2 & 7.9 & 6.8 & 6.1 & 6.5 \end{bmatrix}; \quad \text{sum}(W) = 3.334; \quad D \times W^T \approx \begin{bmatrix} 11.2907 \\ 22.9617 \\ 10.1136 \\ 17.0506 \\ 16.4766 \\ 22.3736 \end{bmatrix};$$

$$D_{real} - D_{combined,1} = \left( \begin{bmatrix} 3.000 \\ 7.000 \\ 3.000 \\ 5.000 \\ 5.000 \\ 7.000 \end{bmatrix} - \begin{bmatrix} 3.3876 \\ 6.8916 \\ 3.0333 \\ 5.1161 \\ 4.9423 \\ 6.7111 \end{bmatrix} \right) = \begin{bmatrix} -0.3876 \\ 0.1084 \\ -0.0333 \\ -0.1161 \\ 0.0577 \\ 0.2889 \end{bmatrix} \quad and \quad 1^T = [1 \quad 1 \quad 1 \quad 1 \quad 1]$$

$$\Rightarrow \quad gradient = -2\, mean\left( \left(D_{real} - D_{combined,k}\right) \frac{sum(W) \times D - \left(D \times W^T\right) \times 1^T}{\left(sum(W)\right)^2} \right)$$

$$= -2\, mean\left( \begin{bmatrix} -0.3876 \\ 0.1084 \\ -0.0333 \\ -0.1161 \\ 0.0577 \\ 0.2889 \end{bmatrix} \left( \frac{3.334 \times \begin{bmatrix} 3.3 & 5.2 & 2.9 & 3.0 & 3.1 \\ 7.5 & 7.0 & 6.9 & 7.2 & 6.8 \\ 3.1 & 3.0 & 3.5 & 2.8 & 2.7 \\ 4.7 & 5.0 & 5.3 & 5.2 & 4.5 \\ 5.9 & 5.3 & 4.8 & 3.9 & 4.7 \\ 7.2 & 7.9 & 6.8 & 6.1 & 6.5 \end{bmatrix} - \begin{bmatrix} 11.2907 \\ 22.9617 \\ 10.1136 \\ 17.0506 \\ 16.4766 \\ 22.3736 \end{bmatrix} \times [1 \quad 1 \quad 1 \quad 1 \quad 1]}{(3.334)^2} \right) \right)$$

$$\Rightarrow \quad gradient \approx [-0.0340 \quad 0.0117 \quad -0.0074 \quad 0.0231 \quad 0.0151]$$

**Step 7** updates Weights based on the calculated *gradient*, where $W_{new} = W - \eta \times gradient$
$= [0.5, 0.517, 0.867, 0.57, 0.88] - 0.1 \times [-0.0340 \quad 0.0117 \quad -0.0074 \quad 0.0231 \quad 0.0151]$
$\Rightarrow W_{new} = [0.5034 \quad 0.5158 \quad 0.8677 \quad 0.5677 \quad 0.8785]$;

**Step 8** ensures $W_{new} > 0$ and $W_{new} = \begin{cases} W_{new} & \text{if } W_{new} > 0 \\ 0 & \text{if } W_{new} < 0 \end{cases} = [0.5034\ 0.5158\ 0.8677\ 0.5677\ 0.8785]$;

**Step 9** normalizes the total weight to 1, where: $sum(W_{new}) = 3.3331$. As a result, $W_{new_normalized} = W_{new}/sum(W_{new}) = [0.5034 \quad 0.5158 \quad 0.8677 \quad 0.5677 \quad 0.8785] \ / \ 3.3331$, therefore $W_{new_normalized} \approx [0.1510 \quad 0.1547 \quad 0.2603 \quad 0.1703 \quad 0.2635]$;

**Step 10** check constraint $\| W_{new} - W_{init} \|_1 > \delta$ to ensure total magnitude of weight changes does not exceed $\delta$. As a result, $\| W_{new} - W_{init} \|_1 \approx 0.0091 < 2$, thus no adjustment needed and move to **Step 13** to update $W$. As a result, $W = W_{new_normalized} \approx [0.140, 0.119, 0.284, 0.171, 0.286]$;

**End of iteration 1** – Output parameters are as follows:

$$D_{combined,1} \approx \begin{bmatrix} 3.3876 \\ 6.8916 \\ 3.0333 \\ 5.1161 \\ 4.9423 \\ 6.7111 \end{bmatrix} \quad \text{and} \quad Error_1 = \begin{bmatrix} 0.1502 \\ 0.0118 \\ 0.0011 \\ 0.0135 \\ 0.0033 \\ 0.0835 \end{bmatrix}; W_{new_normalized} \approx [0.1510 \quad 0.1547 \quad 0.2603 \quad 0.1703 \quad 0.2635].$$

## Data availability

Data will be made available on request.

## References

[1] A. Abu-Siada, S. Hmood, A new fuzzy logic approach to identify power transformer criticality using dissolved gas-in-oil analysis, Int. J. Electr. Power Energy Syst. 67 (2014) 401–408, https://doi.org/10.1016/j.ijepes.2014.12.017.

[2] M.J. Heathcote, Electric Power Transformer Engineering, third ed, CRC Press, Boca Raton, FL, 2012.

[3] V.A. Thiviyanathan, P.J. Ker, Y.S. Leong, F. Abdullah, A. Ismail, M.Z. Jamaludin, Power transformer insulation system: A review on the reactions, fault detection, challenges and future prospects, Alex. Eng. J. 61 (2022) 7697–7713, https://doi.org/10.1016/j.aej.2022.01.026.

[4] H.-C. Sun, Y.-C. Huang, C.-M. Huang, A review of dissolved gas analysis in power transformers, Energy Proc 14 (2012) 1220–1225, https://doi.org/10.1016/j.egypro.2011.12.1079.

[5] S.A. Wani, A.S. Rana, S. Sohail, O. Rahman, S. Parveen, S.A. Khan, Advances in DGA based condition monitoring of transformers: A review, Renew. Sustain. Energy Rev. 149 (2021) 111347, https://doi.org/10.1016/j.rser.2021.111347.

[6] A. Nanfak, I. Fofana, E. Samuel, C.H. Kom, F. Meghnefi, M.G. Ngaleu, Traditional fault diagnosis methods for mineral oil-immersed power transformer based on dissolved gas analysis: past, present and future, IET Nanodielectr 7 (2024) 97–130, https://doi.org/10.1049/nde2.12082.

[7] S. Bustamante, M. Manana, A. Arroyo, A. Laso, R. Martinez, Evolution of graphical methods for the identification of insulation faults in oil-immersed power transformers: A review, Renew. Sustain. Energy Rev. 199 (2024) 114473, https://doi.org/10.1016/j.rser.2024.114473.

[8] A. Wajid, et al., Comparative performance study of dissolved gas analysis (DGA) methods for identification of faults in power transformer, Int. J. Energy Res. (2023) 1–14, https://doi.org/10.1155/2023/9960743.

[9] IEEE guide for the interpretation of gases generated in mineral oil-immersed transformers, IEEE (2019), https://doi.org/10.1109/IEEESTD.2019.8890040.

[10] R. Rogers, IEEE and IEC codes to interpret incipient faults in transformers, using gas in oil analysis, IEEE Trans. Electr. Insul. 13 (1978) 349–354, https://doi.org/10.1109/TEI.1978.298141. EI-.

[11] M. Duval, The duval triangle for load tap changers, non-mineral oils and low temperature faults in transformers, IEEE Electr. Insul. Mag. 24 (2008) 22–29, https://doi.org/10.1109/mei.2008.4665347.

[12] D. Bhalla, R.K. Bansal, H.O. Gupta, Function analysis based rule extraction from artificial neural networks for transformer incipient fault diagnosis, Int. J. Electr. Power Energy Syst. 43 (2012) 1196–1203, https://doi.org/10.1016/j.ijepes.2012.06.042.

[13] S. Seifeddine, B. Khmais, C. Abdelkader, Power transformer fault diagnosis based on dissolved gas analysis by artificial neural network, in: Proc. IEEE Renew. Energies Veh. Technol. (REVET), 2012, pp. 230–236, https://doi.org/10.1109/revet.2012.6195276.

[14] K. Chatterjee, V.K. Jadoun, S. Dawn, R.K. Jarial, Novel prediction-reliability based graphical DGA technique using multi-layer perceptron network & gas ratio combination algorithm, IET Sci. Meas. Technol. 13 (2019) 836–842, https://doi.org/10.1049/iet-smt.2018.5397.

[15] S.S.M. Ghoneim, I.B.M. Taha, N.I. Elkalashy, Integrated ANN-based proactive fault diagnostic scheme for power transformers using dissolved gas analysis, IEEE Trans. Dielectr. Electr. Insul. 23 (2016) 1838–1845, https://doi.org/10.1109/tdei.2016.005301.

[16] L. Tightiz, M.A. Nasab, H. Yang, A. Addeh, An intelligent system based on optimized ANFIS and association rules for power transformer fault diagnosis, ISA Trans 103 (2020) 63–74, https://doi.org/10.1016/j.isatra.2020.03.022.

[17] S.A. Khan, T. Islam, M.D. Equbal, A comprehensive comparative study of DGA based transformer fault diagnosis using fuzzy logic and ANFIS models, IEEE Trans. Dielectr. Electr. Insul. 22 (2015) 590–596, https://doi.org/10.1109/tdei.2014.004478.

[18] T. Kari, et al., An integrated method of ANFIS and Dempster-Shafer theory for fault diagnosis of power transformer, IEEE Trans. Dielectr. Electr. Insul. 25 (2018) 360–371, https://doi.org/10.1109/tdei.2018.006746.

[19] Y. Benmahamed, A. Boubakeur, M. Teguar, Application of SVM and KNN to Duval Pentagon 1 for transformer oil diagnosis, IEEE Trans. Dielectr. Electr. Insul. 24 (2017) 3443–3451, https://doi.org/10.1109/tdei.2017.006841.

[20] L. Hong, Z. Chen, Y. Wang, M. Shahidehpour, M. Wu, A novel SVM-based decision framework considering feature distribution for Power Transformer fault Diagnosis, Energy Rep 8 (2022) 9392–9401, https://doi.org/10.1016/j.egyr.2022.07.062.

[21] B. Zeng, W. Zhu, J. Guo, S. Huang, Z. Xiao, F. Yuan, A transformer fault diagnosis model based on hybrid grey wolf optimizer and LS-SVM, Energies 12 (2019) 4170, https://doi.org/10.3390/en12214170.

[22] K.N.V.P.S. Rajesh, I. Fofana, U.M. Rao, P. Rozga, A. Paramane, Influence of data balancing on transformer DGA fault classification with machine learning algorithms, IEEE Trans. Dielectr. Electr. Insul. 30 (2023) 385–392, https://doi.org/10.1109/TDEI.2022.3230377.

[23] P.A.R. Azmi, M. Yusoff, M.T.M. Sallehud-Din, Improving transformer failure classification on imbalanced DGA data using data-level techniques and machine learning, Energy Rep 13 (2025) 264–277, https://doi.org/10.1016/j.egyr.2024.12.006.

[24] A. Nanfak, C.H. Kom, I. Fofana, F. Meghnefi, S. Eke, G.M. Ngaleu, Hybrid DGA method for power transformer faults diagnosis based on evolutionary k-means clustering and dissolved gas subsets analysis, IEEE Trans. Dielectr. Electr. Insul. 30 (2023) 2421–2428, https://doi.org/10.1109/TDEI.2023.3275119.

[25] I.B.M. Taha, S. Ibrahim, D.-E.A. Mansour, Power transformer fault diagnosis based on DGA using a convolutional neural network with noise in measurements, IEEE Access 9 (2021) 111162–111170, https://doi.org/10.1109/access.2021.3102415.

[26] L. Jin, A. Abu-Siada, D. Kim, K.Y. Chan, Deep machine learning-based asset management approach for oil-immersed power transformers using dissolved gas analysis, IEEE Access 12 (2024) 27794–27809, https://doi.org/10.1109/ACCESS.2024.3366905.

[27] S.S.M. Ghoneim, M. Lehtonen, K. Mahmoud, M.M.F. Darwish, Enhancing diagnostic accuracy of transformer faults using teaching-learning-based optimization, IEEE Access 9 (2021) 30817–30832, https://doi.org/10.1109/access.2021.3060288.

[28] D. Zou, et al., Transformer fault classification for diagnosis based on DGA and deep belief network, Energy Rep 9 (2023) 250–256, https://doi.org/10.1016/j.egyr.2023.09.183.

[29] H. Malik, R. Sharma, S. Mishra, Fuzzy reinforcement learning based intelligent classifier for power transformer faults, ISA Trans 101 (2020) 390–398, https://doi.org/10.1016/j.isatra.2020.01.016.

[30] H. He, J. Yu, W.-J. Lee, D. Luo, W. Liang, A power transformer fault diagnosis method based on improved variational quantum shadow learning, IEEE Trans. Power Deliv. (2025) 1–13, https://doi.org/10.1109/TPWRD.2025.3546722.

[31] K.A. Nguyen, H.H. Le, B.T. Phung, H.V. Tran, A fuzzy logic approach combined with IEC and Roger methods for power transformer fault diagnosis based on DGA, Univ. Danang J. Sci. Technol. 23 (2025) in press.

[32] K.A. Nguyen, H.H. Le, B.T. Phung, H.V. Tran, Enhancing data preprocessing layer for power transformer fault diagnosis based on dissolved gas analysis combining fuzzy logic and Duval triangle 1, in: Proc. 8th Int. Conf. Circuits Syst. Simul. (ICCSS 2025), Ind. Univ. Ho Chi Minh City, Vietnam, 2025, in press.

[33] Y.-C. Huang, H.-C. Sun, Dissolved gas analysis of mineral oil for power transformer fault diagnosis using fuzzy logic, IEEE Trans. Dielectr. Electr. Insul. 20 (2013) 974–981, https://doi.org/10.1109/TDEI.2013.6518967.

[34] M. Arshad, S. Islam, A. Khaliq, Fuzzy logic approach in power transformers management and decision making, IEEE Trans. Dielectr. Electr. Insul. 21 (2014) 2343–2354, https://doi.org/10.1109/TDEI.2014.003859.

[35] R.A. Prasojo, B.A. Soedjarno, H. Gumilang, N.U. Maulidevi, S. Suwarno, A fuzzy logic model for power transformer faults' severity determination based on gas level, gas rate, and dissolved gas analysis interpretation, Energies 13 (2020) 1009, https://doi.org/10.3390/en13041009.

[36] B.C. Mateus, J.T. Farinha, M. Mendes, Fault detection and prediction for power transformers using fuzzy logic and neural networks, Energies 17 (2024) 296, https://doi.org/10.3390/en17020296.

[37] S.Q. Su, P. Austin, L.L. Lai, C. Mi, A fuzzy dissolved gas analysis method for the diagnosis of multiple incipient faults in a transformer, IEEE Trans. Power Syst. 15 (2000) 593–598, https://doi.org/10.1109/59.867146.

[38] I.B.M. Taha, S.S.M. Ghoneim, A. Hoballah, Optimal ratio limits of rogers' four-ratios and IEC 60599 code methods using particle swarm optimization fuzzy-logic approach, IEEE Trans. Dielectr. Electr. Insul. 27 (2020) 222–230, https://doi.org/10.1109/TDEI.2019.008395.

[39] I.B.M. Taha, S.S.M. Ghoneim, H.G. Zaini, A fuzzy diagnostic system for incipient transformer faults based on DGA of the insulating transformer oils, Int. Rev. Electr. Eng. 11 (2016) 305, https://doi.org/10.15866/iree.v11i3.8453.

[40] S.A. Wani, D. Gupta, G. Prashal, S.A. Khan, Smart diagnosis of incipient faults using dissolved gas analysis-based fault interpretation matrix (FIM), Arab, J. Sci. Eng. 44 (2019) 6977–6985, https://doi.org/10.1007/s13369-019-03739-4.

[41] S. Hmood, S.M. Islam, A. Abu-Siada, M.A.S. Masoum, Standardization of DGA interpretation techniques using fuzzy logic approach, in: Proc. 2012 Int. Conf. Condition Monitoring Diagnosis, 2012, pp. 929–932, https://doi.org/10.1109/CMD.2012.6416305.

[42] G.K. Irungu, J.L. Munda, A.O. Akumu, A new fault diagnostic technique in oil-filled electrical equipment; the dual of Duval triangle, IEEE Trans. Dielectr. Electr. Insul. 23 (2016) 3405–3410, https://doi.org/10.1109/TDEI.2016.005927.

[43] S.A. Wani, M.U. Farooque, S.A. Khan, D. Gupta, Multiple incipient fault classification approach for enhancing the accuracy of dissolved gas analysis (DGA), IET Sci. Meas. Technol. 13 (2019) 959–967, https://doi.org/10.1049/iet-smt.2018.5135.

[44] N. Poonnoy, T. Suwanasri, C. Suwanasri, Fuzzy logic approach to dissolved gas analysis for power transformer failure index and fault identification, Energies 14 (2020) 36, https://doi.org/10.3390/en14010036.

[45] M. Arshad, S. Islam, Significance of cellulose power transformer condition assessment, IEEE Trans. Dielectr. Electr. Insul. 18 (2011) 1591–1598, https://doi.org/10.1109/TDEI.2011.6032829.

[46] L.A. Zadeh, Outline of a new approach to the analysis of complex systems and decision processes, IEEE Trans. Syst. Man Cybern. 3 (1973) 28–44, https://doi.org/10.1109/TSMC.1973.5408575. SMC-.

[47] X. Dong, D.-X. Zhou, Learning gradients by a gradient descent algorithm, J. Math. Anal. Appl. 341 (2007) 1018–1027, https://doi.org/10.1016/j.jmaa.2007.10.044.

[48] J. Cai, H. Wang, D.-X. Zhou, Gradient learning in a classification setting by gradient descent, J. Approx. Theory 161 (2008) 674–692, https://doi.org/10.1016/j.jat.2008.12.002.

[49] P.C. Jackson, Introduction to artificial intelligence, ACM SIGART Bull 47 (1974) 15, https://doi.org/10.1145/1045190.1045191.

[50] N.V. Nga, N.H. Chien, D. Truc, T.D. Tho, N.V. Luc, T.H. Vu, Research on application of artificial intelligence in diagnosis of potential failures in transformers by dissolved gas analysis method, Univ. Danang J. Sci. Technol. 22 (2024) 30–35. https://jst-ud.vn/jst-ud/article/view/9362.

[51] E. Li, L. Wang, B. Song, Fault diagnosis of power transformers with membership degree, IEEE Access 7 (2019) 28791–28798, https://doi.org/10.1109/ACCESS.2019.2902299.

[52] S.S.M. Ghoneim, I.B.M. Taha, A new approach of DGA interpretation technique for transformer fault diagnosis, Int. J. Electr. Power Energy Syst. 81 (2016) 265–274, https://doi.org/10.1016/j.ijepes.2016.02.018.
[53] O.E. Gouda, S.H. El-Hoshy, H.H. El-Tamaly, Proposed three ratios technique for the interpretation of mineral oil transformers based dissolved gas analysis, IET Gener. Transm. Distrib. 12 (2018) 2650–2661, https://doi.org/10.1049/iet-gtd.2017.1927.
[54] M. Duval, A. Depabla, Interpretation of gas-in-oil analysis using new IEC publication 60599 and IEC TC 10 databases, IEEE Electr. Insul. Mag. 17 (2001) 31–41, https://doi.org/10.1109/57.917529.
[55] S. Suwarno, H. Sutikno, R.A. Prasojo, A. Abu-Siada, Machine learning based multi-method interpretation to enhance dissolved gas analysis for power transformer fault diagnosis, Heliyon 10 (2024) e25975, https://doi.org/10.1016/j.heliyon.2024.e25975.

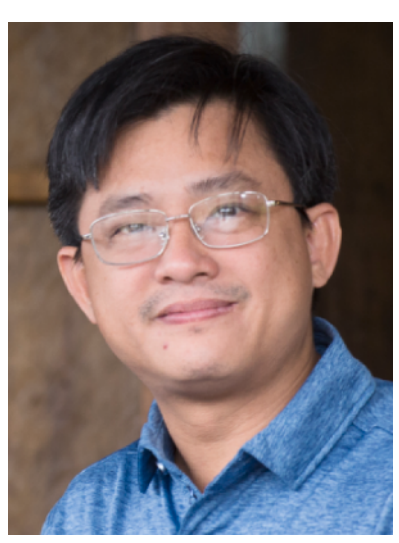

**Kim-Anh Nguyen** is currently a lecturer at The University of Danang - University of Science and Technology, Vietnam. He served as the Head of the Automation Division in the Faculty of Electrical Engineering from 2016 to 2020. He received his Engineer's degree in Electrical Engineering from the University of Science and Technology in 2004, an M.Eng. degree in Control Engineering and Automation from the University of Danang in 2009, and a Ph.D. in Systems Optimization and Dependability from Troyes University of Technology, France, in 2015. His current research interests include stochastic modeling of system deterioration, diagnosis, prognosis and health management techniques, maintenance decision-making, reliability engineering, power electronics, power quality, SCADA, IoT, fuzzy logic, artificial intelligence, solar power system, embedded system, and control systems.

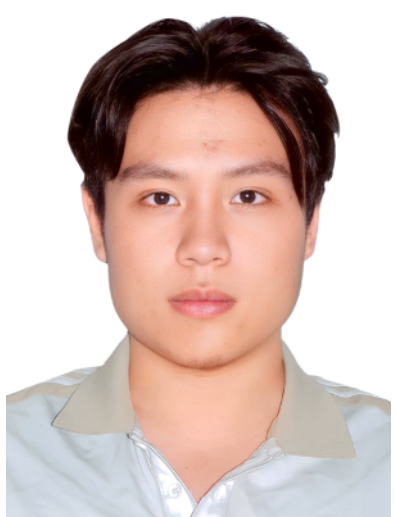

**Huy Hoang Le** is currently an undergraduate student in Control and Automation Engineering at The University of Danang - University of Science and Technology, Vietnam. His academic interests include control theory, artificial intelligence, fuzzy systems, machine learning, quantum computing, industrial IoT, SCADA applications and embedded system design. He contributes to the development of intelligent and adaptive control systems, with hands-on experience in system modeling and algorithm implementation using MATLAB/Simulink, as well as the application of quantum algorithms through various quantum simulation platforms.

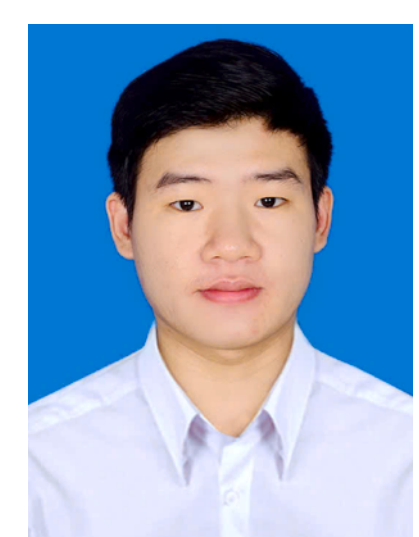

**Ba Tu Phung** is currently an undergraduate student majoring in Control and Automation Engineering at The University of Danang - University of Science and Technology, Vietnam. His current research interests include control systems, artificial intelligence, fuzzy logic, fault diagnosis, IoT, SCADA systems, PLC programming, and embedded systems for industrial automation and intelligent monitoring. His work focuses on applying AI techniques to enhance system reliability and decision-making, combined with hands-on experience in modeling, simulation, and implementation using MATLAB/Simulink and Python.